\documentclass[letterpaper]{article} 
\usepackage{aaai25}  
\usepackage{times}  
\usepackage{helvet}  
\usepackage{courier}  
\usepackage[hyphens]{url}  
\usepackage{graphicx} 
\usepackage{natbib}  
\usepackage{caption} 
\usepackage{amsfonts}
\usepackage{amsmath}
\usepackage{cleveref}
\usepackage{multirow}
\usepackage{xcolor}
\usepackage{colortbl}
\usepackage{lipsum}
\usepackage{verbatim}
\usepackage{newfloat}
\usepackage{listings}
\DeclareCaptionStyle{ruled}{labelfont=normalfont,labelsep=colon,strut=off} 
\nocopyright 

\title{CodecNeRF: Toward Fast Encoding and Decoding, Compact, and High-quality Novel-view Synthesis}
\author{
    Gyeongjin Kang\textsuperscript{\rm 1}\equalcontrib, 
    Younggeun Lee\textsuperscript{\rm 2}\equalcontrib, 
    Seungjun Oh\textsuperscript{\rm 2}, 
    Eunbyung Park\textsuperscript{\rm 1, \rm2}\thanks{Corresponding author}
}

\affiliations{
    \textsuperscript{\rm 1}Department of Electrical and Computer Engineering, Sungkyunkwan University\\
    \textsuperscript{\rm 2}Department of Artificial Intelligence, Sungkyunkwan University
}

\begin{document}

\maketitle

\begin{abstract}
Neural Radiance Fields (NeRF) have achieved huge success in effectively capturing and representing 3D objects and scenes. However, to establish a ubiquitous presence in everyday media formats, such as images and videos, we need to fulfill three key objectives: 1. fast encoding and decoding time, 2. compact model sizes, and 3. high-quality renderings. Despite recent advancements, a comprehensive algorithm that adequately addresses all objectives has yet to be fully realized. In this work, we present CodecNeRF, a neural codec for NeRF representations, consisting of an encoder and decoder architecture that can generate a NeRF representation in a single forward pass. Furthermore, inspired by the recent parameter-efficient finetuning approaches, we propose a finetuning method to efficiently adapt the generated NeRF representations to a new test instance, leading to high-quality image renderings and compact code sizes. The proposed CodecNeRF, a newly suggested encoding-decoding-finetuning pipeline for NeRF, achieved unprecedented compression performance of more than 100$\times$ and remarkable reduction in encoding time while maintaining (or improving) the image quality on widely used 3D object datasets.
\noindent
{\color{black} \rule{\linewidth}{0.2mm}}
Project page: \url{https://gynjn.github.io/CodecNeRF}
\end{abstract}


%

\section{Introduction}
\label{sec:intro}

Neural Radiance Fields (NeRF) have been enormously successful in representing 3D scenes~\cite{nerf}. Given a handful of pictures taken from various viewpoints, it generates photo-realistic images from novel viewpoints, proving beneficial for various applications, such as 3D photography and navigation~\cite{kuang2022neroic, jampani2021nerf_3dphotography, kuang2023palettenerf, adamkiewicz2022nerf_navigation, kwon2023renderablenavigation, maggio2023locnerf}. In addition, ongoing research endeavors have enhanced its compatibility with conventional graphics rendering engines by enabling mesh and texture extraction~\cite{Munkberg2022nerfmesh, rakotosaona2023nerfmeshing,baatz2022nerftexture, tang2023nerf2mesh, munkberg2022extractingnvidia}, and thus, it further expands its usability.
Moreover, the recent 3D generation and editing techniques make it more valuable as a next-generation 3D media representation, opening new possibilities and applications.

The primary reason contributing to the longstanding success of image and video is the widespread adoption of standard codec software and hardware~\cite{bross202H266,sullivan2012H265, wiegand2003H264, rijkse1996H263}. We simply take a picture or video with our hand-held devices, and the encoder rapidly compresses the data. Then, the encoded data are transmitted over network communication channels, and the receivers can consume the data with the help of fast decoding software and hardware. We envision similar usage of 3D media using NeRF: 1) senders obtain multi-view images, 2) an encoder turns those images into a NeRF representation (encoding), 3) the encoded representation is communicated through the network, 4) receivers decode the encoded data and users enjoy the contents by rendering from various viewpoints. We urge the development of an algorithmic pipeline that can achieve rapid encoding and decoding speeds, compact data sizes, and high-quality view synthesis to support this common practice.

Despite considerable technological progress, there has yet to be a fully satisfying solution to achieve all of the stated goals. Training speed (encoding time) has remarkably advanced from days to a few hours or minutes~\cite{chen2022tensorf, kplanes_2023, kerbl20233dgs, muller2022instantngp, takikawa2023compactngp, sun2022dvgo, fridovich2022plenoxels, liu2020nsvf}. However, due to the inherent drawback of the per-scene optimization approach, they still require powerful GPU devices and at least tens of thousands of training iterations to converge. The encoder-decoder approaches, which generate NeRF in a single network forward pass, have been proposed~\cite{wang2021ibrnet, trevithick2021grf, chen2021mvsnerf, lin2023visionnerf, yu2021pixelnerf, li2021mine, johari2022geonerf, liu2022neuralrays, dupont2020equivariant, chibane2021srf, raj2021pva, rematas2021sharf}. However, they primarily focus on few-shot generalization and do not consider the codec aspects, and the rendering image quality is limited compared to the optimization-based approaches. On the other hand, there has been extensive investigation into compact NeRF representations to minimize the encoded data sizes~\cite{takikawa2023compactngp, rho2023maskedwavelet, takikawa2022variablebitrate, li2023compressing1mb, shin2024binarynerf, tang2022compressible, bird20213dcompressionentropy, lu2021compressiveneuralrepresentation, lee2023compact3dgs, deng2023compressingexplicit}. While successful, the suggested methods are mostly based on the per-scene optimization approach, resulting in longer training iterations.

In this work, we introduce CodecNeRF, a neural codec for NeRF designed to accomplish the previously mentioned objectives all at once. The proposed neural codec consists of a novel encoder and decoder architectures that can produce a NeRF representation in a single forward pass. The encoder takes multi-view images and produces compact codes that are transmitted to other parties through network communications. The decoder that is present on both the sender and receiver sides generates the NeRF representations given the delivered codes. This forward-pass-only approach, as demonstrated numerous times by preceding neural codecs for image and video, can achieve rapid encoding/decoding times and exceptional compression performance.

The forward pass alone, however, does not guarantee that the generated NeRF representation synthesizes high-quality images. The primary issue stems from the scarcity of instances and diversities of the existing 3D datasets, in contrast to the abundance found in image and video domains. This shortage hampers the trained models' capability to effectively generalize to new 3D test instances. Therefore, we propose to finetune the NeRF representations on the sender side and further transmit the finetuned `delta' information to the receiver along with the codes. Then, the decoder on the receiver side uses the transmitted codes to reproduce the initial NeRF representations and apply `delta' to obtain the final NeRF representations. Since the initial NeRF representations from the forward pass are already well-formed, the subsequent finetuning requires far fewer iterations than the per-scene optimization approach, which results in significantly faster encoding time.

*To reduce the overall size of the final code (codes $+$ finetuning `delta'), we suggest parameter-efficient finetuning (PEFT) techniques on the initial NeRF representations~\cite{hu2021lora}. Finetuning the entire decoder or NeRF representations substantially increases the code sizes to be transmitted, negating the advantages of employing the encoder and decoder methodology. In this work, the NeRF representation is based on K-planes method consisting of multi-resolution plane features and an MLP network. We employ the widely used low-rank adaptation (LoRA) methods for the MLP and suggest a novel PEFT technique for plane features inspired by the low-rank tensor decomposition method.

We have conducted comprehensive experiments using two representative 3D datasets, Objaverse~\cite{deitke2023objaverse, deitke2024objaversexl} and Google Scanned Objects~\cite{downs2022google}. The experimental results show that the proposed encoder-decoder-finetuning method, CodecNeRF, achieved 100$\times$ more compression performance and 
significantly reduced encoding (training) time over the per-scene optimization baseline method (triplane) while maintaining the rendered image quality. Additionally, we evaluated the  CodecNeRF's performance on real scenes using the DTU dataset~\cite{jensen2014dtu}. We perceive this outcome as unlocking new research opportunities and application avenues using NeRF. 
The main contributions can be summarized as follows:

\begin{itemize}
      \item We propose CodecNeRF, an encoder-decoder-finetuning pipeline for the newly emerging NeRF representation.
      \item We design novel 3D-aware encoder-decoder architectures, efficiently aggregating multi-view images, generating compact codes, and making NeRF representations from the codes.
      \item We present the parameter-efficient finetuning approach for further finetuning the NeRF representations that consist of MLP and feature planes.
      \item We achieved the unprecedented compression ratio and encoding speedup of NeRF while preserving high-quality rendering.
\end{itemize}

\section{Related Works}
\label{sec:related_works}

\noindent\textbf{Fast training NeRF.}
To reduce the computational complexity, grid-based representations have been suggested as an alternative to MLP.
Plenoxels~\cite{fridovich2022plenoxels} constructed a sparse voxel grid with density and color value at each voxel explicitly. DVGO~\cite{sun2022dvgo} and Instant-NGP~\cite{muller2022instantngp} utilized voxel grids that store features and densities and employed a tiny MLP to compute the final output values.
TensoRF~\cite{chen2022tensorf} decomposed 3D grids in an axis-aligned manner via VM decomposition and CP decomposition for further improving the parameter efficiency. Inspired by EG3D~\cite{chan2022eg3d}, K-Planes~\cite{kplanes_2023} employed multi-scale orthogonal 2D planes, triplanes, showing scalability to higher dimensions while maintaining the speed advantage of grid-based representations. However, those per-scene optimization methods require numerous iterations to achieve high-quality novel view synthesis. In this work, we propose an encoder-decoder architecture to encode NeRF in a single-forward pass from multi-view input images. Furthermore, we utilize a low-rank decomposition scheme for efficient finetuning, showing fast convergence with few iterations.


\noindent\textbf{Compact NeRF.}
Follow-up studies of NeRF aim to reduce storage size while preserving the performance of the original models. TensorRF~\cite{chen2022tensorf} and CCNeRF~\cite{tang2022compressible} used tensor decomposition and low-rank approximation to reduce model size. Related to quantization methods, VQRF ~\cite{li2023compressing1mb} introduced trainable vector quantization method and VBNF ~\cite{takikawa2022variablebitrate} compressed feature grid by employing vector-quantized auto-decoder. Masked wavelet representation~\cite{rho2023maskedwavelet} applied wavelet transform on grid-based NeRF and quantization on coefficients with trainable mask and BiRF~\cite{shin2024binarynerf} proposed binary-based radiance field that quantizes each feature with binary values. Recently, NeRFCodec~\cite{li2024nerfcodec} achieved a high compression ratio by combining pretrained neural image codec and entropy coding. However, it requires per-scene optimization process to obtain the NeRF representations, which demands significant encoding time. Compared to the aforementioned methods, our model is a forward-pass-based approach that achieves fast encoding of 3D representations.


\begin{figure*}[t]
  \centering
  \includegraphics[width=0.90\linewidth]{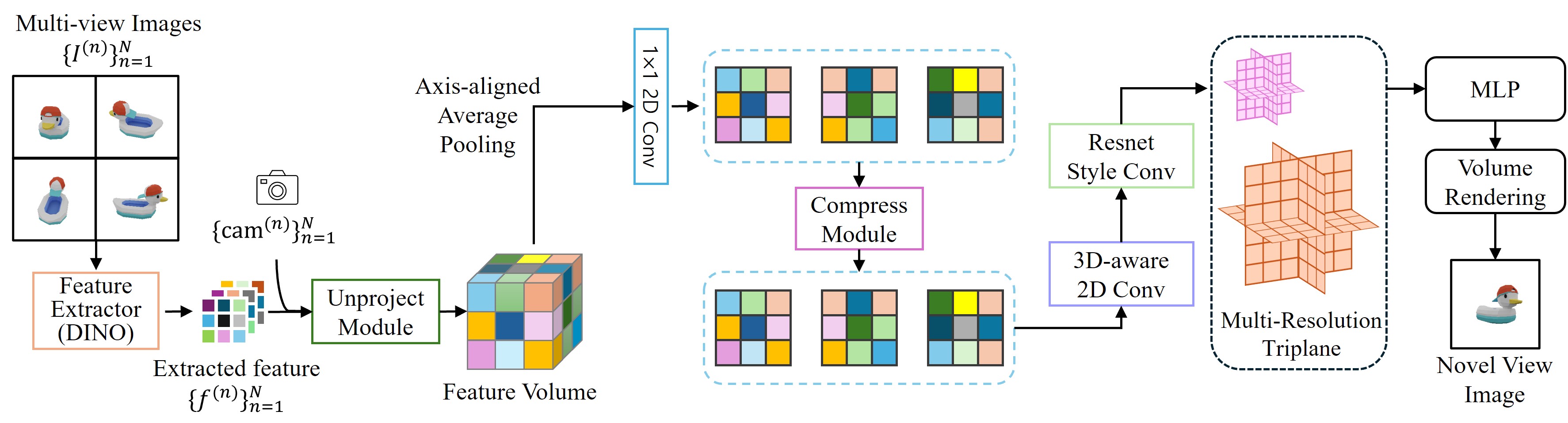}
  \caption{CodecNeRF encoder and decoder architecture.} 
  \label{fig:overall_arch}
\end{figure*}

\noindent\textbf{Neural codec for images and videos.}
A large body of works have explored the application of learning-based methods for compressing various types of data.
In the image domain, along with CNN's remarkable property as a feature extractor, encoder-decoder~\cite{baldi2012ae, kingma2013vae} based methods proposed by Ball\'e~\cite{balle2016compiclr17, balle2018compiclr18} are established as standard approaches. These models are combined with an entropy coding, such as ~\cite{rissanen1979arithmetic, martin1979range, huffman1952method}, and trained to minimize the discretized code length while weighing the trade-offs between bit-rate and representation distortion. Learning-based video compression methods, expanded from image techniques, have incorporated time axis using optical flow~\cite{lu2019dvc}, reference frames~\cite{lin2020mlvc}, and contextual learning~\cite{li2021dcvc, sheng2022temporal}. Inspired by neural compression methods in images and videos, we propose CodecNeRF, the first encoder-decoder based learned codec for NeRF, integrating neural codec with parameter-efficient fintuning in a novel way.


\section{CodecNeRF}
This section describes the proposed CodeNeRF pipeline with detailed architectures and finetuning methods. We explain the overall architecture 
(Sec.~\ref{sec:overall_arch}) 
first and present detailed methods in the following sections for each module: 3D feature construction 
(Sec.~\ref{sec:3d_feature}), 3D feature compression 
(Sec.~\ref{sec:3d_feature_compression}), and multi-resolution triplanes 
(Sec.~\ref{sec:multires_triplane}). Then, we present the training objectives used to train the proposed architecture 
(Sec.~\ref{sec:training_loss})
and the parameter-efficient finetuning method for generating compact codes (Sec.~\ref{sec:peft} and \ref{sec:peftec}).

\subsection{Overall architecture}
\label{sec:overall_arch}
Fig.~\ref{fig:overall_arch} depicts the overall encoder and decoder architecture of CodecNeRF. Given $N$ input images from different viewpoints, $\{I^{(n)}\}_{n=1}^N$, the goal is to produce a NeRF representation (multi-resolution triplanes).
First, a 2D image feature extractor module, $\texttt{feat}_\theta$, processes all input images and generates 2D feature maps for each input image, $\{f^{(n)}\}_{n=1}^N$. 
Then, the unproject and aggregation module, $\texttt{unproj}$ and $\texttt{agg}_\phi$, lifts the 2D features to 3D features and aggregates the unprojected 3D features into a single 3D feature, $f_{3D} \in \mathbb{R}^{C \times V  \times V  \times V}$ (to avoid clutter notation, we assume height, width, and depth resolutions are same, $V$). The 3D feature, $f_{3D}$, is further processed by axis-aligned average pooling along each axis, resulting in three 2D features (a 2D feature for each axis), $f_{xy}, f_{yz}, f_{xz} \in \mathbb{R}^{C \times V  \times V}$. These 2D features are used to generate multi-resolution triplanes by $\texttt{triplane}_\psi$, and finally, we perform the volumetric rendering to render an image using $\texttt{MLP}_\omega$. Furthermore, 2D features $f_{xy},f_{yz}$, and $f_{xz}$ are compressed by the compression module, $\texttt{comp}_\chi$, producing the minimal sizes of the codes to be transmitted. The entire pipeline is differentiable, and we train end-to-end to optimize all learnable parameters, $\{\theta, \phi, \chi, \psi, \omega \}$.


\subsection{3D feature from multi-view images}
\label{sec:3d_feature}
In this submodule, we construct the 3D feature from multi-view input images.
To extract 2D image features, we adopt a pre-trained visual transformer (ViT) ~\cite{dosovitskiy2020image}, specifically, DINO~\cite{caron2021dino}
to produce an image features $f^{(n)}$ given an input image $I^{(n)}$.
We process each view image individually using the shared feature extractor, $\texttt{feat}_\theta$.
Following the conventional NeRF training scheme, we assume that we can obtain camera poses for input view images beforehand.
The 3D feature construction can be written as follows.

\begin{equation}
f_{3D} = \texttt{agg}_\phi(\{ \texttt{unproj}(f^{(n)}, \text{cam}^{(n)}, \text{coord}) \}_{n=1}^N),
\end{equation}

where $\text{cam}^{(n)}$ denotes the camera pose for the input views.
Inspired by the recent unprojection methods~\cite{liu2023syncdreamer, Liu202312345}, we first construct a 3D coordinate tensor, $\text{coord} \in \mathbb{R}^{3 \times V \times V \times V}$, whose resolution is $V$ for all axis.
Then, each coordinate is projected into 2D space given the camera pose, and the feature is extracted from the image feature $f^{(n)}$ using bilinear interpolation, generating the intermediate 3D feature.
We use an aggregation module $\texttt{agg}_\phi$ paramaterized $\phi$ to combine $N$ intermediate 3D features and produce the final 3D feature, $f_{3D} \in \mathbb{R}^{C \times V \times V \times V}$.
We use a few 3D convolution layers to aggregate features and further extract useful information.


\subsection{3D feature compression}
\label{sec:3d_feature_compression}

The goal of 3D feature compression is to minimize the number of bits required to reconstruct the final NeRF representations, and $f_{3D}$ from the previous stage is 3D volume, thus inefficient for storage and transmission purposes. In this work, we opt to use explicit-implicit hybrid NeRF representation, triplane~\cite{chan2022eg3d}. Triplane representation decomposes a 3D volume into three 2D planes, serving as a prevalent technique for the NeRF representations~\cite{kplanes_2023, Hexplane, shue20223nfd}. It scales with $O(V^2)$ for the resolution $V$ as opposed to $O(V^3)$ for a dense 3D volume.

We first transform the 3D feature into three 2D features by average pooling along each axis.
\begin{equation}
f_{xy}\hspace{-0.3em}=\hspace{-0.3em}\texttt{ap-z}(f_{3D}), \hspace{-0.1em}f_{yz}\hspace{-0.3em}=\hspace{-0.3em}\texttt{ap-x}(f_{3D}), \hspace{-0.1em}f_{xz}\hspace{-0.3em}=\hspace{-0.3em}\texttt{ap-y}(f_{3D}),
\end{equation}
where $\texttt{ap-x}$ means average pooling along $x$ axis.
Then, $\texttt{comp}_\chi$ compresses the three 2D feature maps using vector quantization methods~\cite{vq, vq-vae}.
It consists of a downsampling CNN, an upsampling CNN, and a codebook ($\chi$ includes all parameters in these three modules).
First the downsampling 2D CNN module process each 2D feature map to generate low-resolution 2D feature map, $l_{xy},l_{yz},l_{xz} \in \mathbb{R}^{C' \times V' \times V'}$ ($C' << C$ and $V' << V$). Then, we find the closest code from the codebook to perform the vector quantization.

\begin{equation}
\label{eq:vq}
\bar{l}_{xy,i,j} = e_{\underset{k}{\operatorname{argmin}} \left\| l_{xy,i,j} - e_k \right\|_2},
\end{equation}
where $e \in \mathbb{R}^{K \times C'}$ is the codebook, $K$ is the codebook size, $e_k \in \mathbb{R}^{C'}$ denotes the $k$-th element of the codebook, $l_{xy,i,j} \in \mathbb{R}^{C'}$ denotes the element of $l_{xy}$ indexed by $(i,j)$ location, and $\bar{l}_{xy,i,j}$ is the vector quantized 2D feature map.
During training, we optimize the codebook $e$, and the loss function for a training instance can be written as,
\begin{equation}
\mathcal{L}_{\mathrm{vq}}=\left\|\text{sg}\left[l \right]-\bar{l} \right\|_2^2+\lambda_\text{commit}\left\|\text{sg}[\bar{l}] - l \right\|_2^2,
\end{equation}
where $\text{sg}\left[\cdot\right]$ is the stop-gradient operator, and $\lambda_\text{commit}$ regulate the commitment to codebook embedding. With the slight abuse of notation, here we define $l, \bar{l} \in \mathbb{R}^{3 \times C' \times V' \times V'}$ as the concatenated tensor of three low-resolution 2D feature maps.
Finally, the upsampling CNN produces three 2D feature maps with the increased resolutions, $\bar{f}_{xy},\bar{f}_{yz},\bar{f}_{xz} \in \mathbb{R}^{C \times V \times V}$. During training, the input to the upsampling CNN is $l$, but $\bar{l}$ is used during testing.

\subsection{Multi-resolution triplanes}
\label{sec:multires_triplane}
The previous works ~\cite{muller2022instantngp, kplanes_2023, lindell2022bacon, kuznetsov2021neumip, hu2023trimiprf, nam2024mipgrid} have shown that using a multi-resolution representation efficiently encodes spatial features at different scales.
It encourages spatial smoothness across different scales, superior convergence, and better accuracy.
Building upon these observations, we propose a hierarchical 3D-aware convolution block, $\texttt{triplane}_\psi$, that generates a multi-resolution triplanes revised from the one introduced in ~\cite{wang2022rodin, wu2023sin3dm}.
It introduces rolled-out triplanes that attend to all components from the relevant rows and columns, enabling cross-plane feature interaction.
\begin{equation}
(\tilde{f}_{xy}, \tilde{f}_{yz}, \tilde{f}_{xz}) =\texttt{triplane}_\psi({\bar{f}_{xy},\bar{f}_{yz},\bar{f}_{xz}}),
\end{equation}
where $\tilde{f}_{xy} =\{ \tilde{f}_{xy}^{1}, \tilde{f}_{xy}^{2} \}$ is a set of multi-resolution triplane features for `$xy$' plane, and $\tilde{f}_{xy}^1 \in \mathbb{R}^{C \times V_{1} \times V_{1}}$ and $\tilde{f}_{xy}^2 \in \mathbb{R}^{C \times V_{2} \times V_{2}}$ are different resolution features.


The proposed triplane renderer consists of two distinct MLP heads, coarse and fine, for decoding the RGBs and densities. Given a 3D coordinate $\mathit{p} \in \mathbb{R}^{3}$, the decoder collects the triplane features at three axis-aligned projected locations of $\mathit{p}_{xy}, \mathit{p}_{yz}, \mathit{p}_{xz} \in \mathbb{R}^2$, using bilinear interpolation. We simply concat the triplane features across the different scales and aggregate by summation.
\begin{align}
&f_\text{tri}(p)=\hspace{-1.7em}\sum_{k \in \{xy,yz,xz\}}\hspace{-1.7em} \texttt{concat}(\texttt{itp}(\tilde{f}_{k}^1, {p}_{k}), \texttt{itp}(\tilde{f}_{k}^2, {p}_{k}) ), \\ &c(p,d), \sigma(p) = \texttt{MLP}_\omega(f_\text{tri}(p), p, \texttt{PE}(d)),
\end{align}
where $\texttt{itp}(\cdot,\cdot)$ bilinearly interpolates the features given the projected 2D coordinates, $\texttt{concat}(\cdot)$ concatenate the interpolated features, $f_\text{tri}(p) \in \mathbb{R}^{3C}$ is the feature to be processed by an MLP network to generate $c(p)$ and $\sigma(p)$, the color and density of a point. $\texttt{PE}(d)$ is the view direction after applying the positional encoding. Finally, the volume rendering ~\cite{max1995volumerendering} is applied to render the images using the two-pass hierarchical importance sampling method proposed by NeRF~\cite{nerf}.

\begin{figure}
  \centering
  \includegraphics[width=\linewidth]{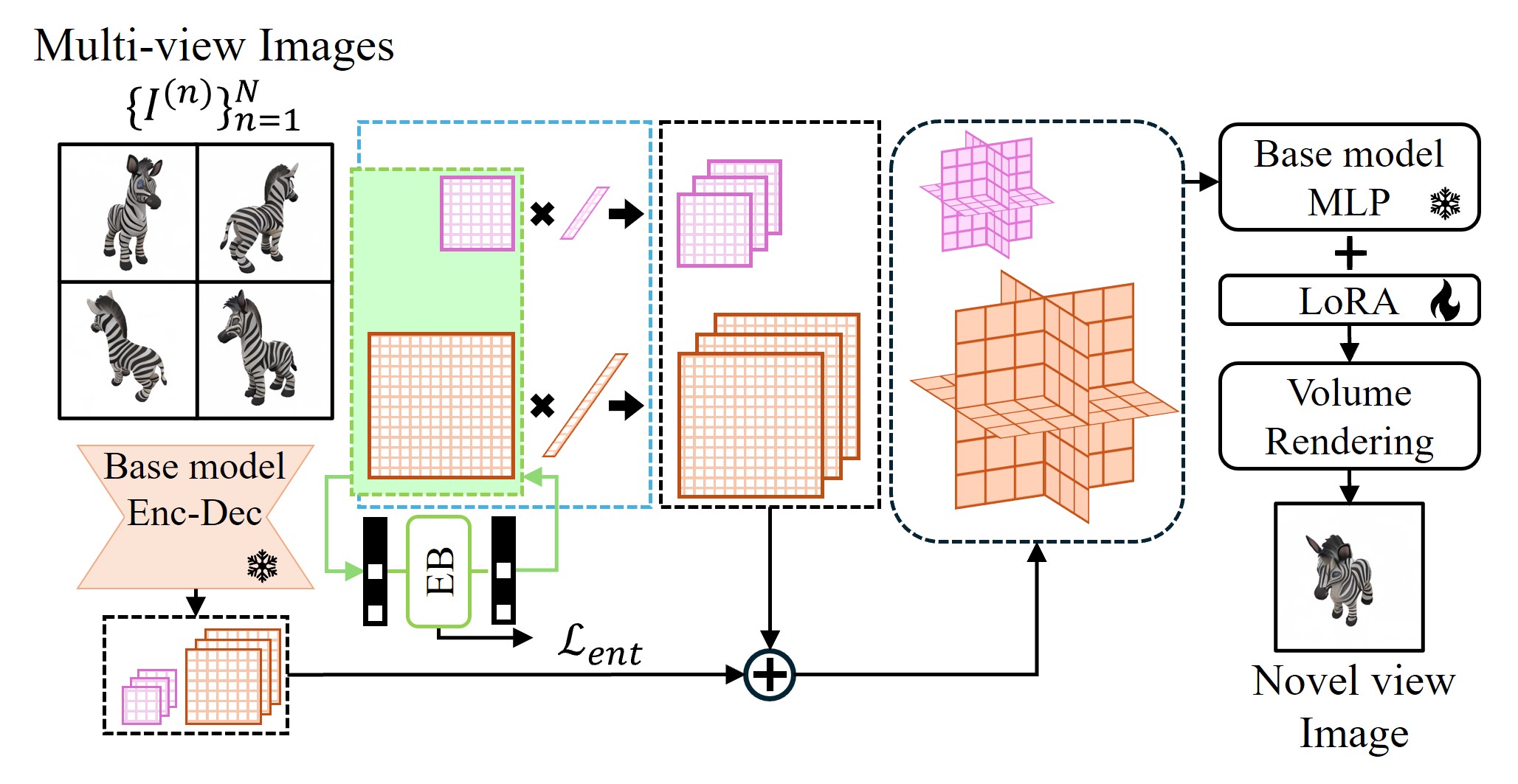}
  \caption{Parameter-efficient finetuning process.} 
  \label{fig:PEFT}
\end{figure}

\begin{table*}[]
\renewcommand{\arraystretch}{1.00}
\resizebox{\linewidth}{!}{%
\begin{tabular}{l|l|ccccccccccccccc} \hline
\multicolumn{1}{c|}{}                       & \multicolumn{1}{c|}{}                         & \multicolumn{14}{c}{Iterations}                         \\ \cline{3-17} 
\multicolumn{1}{c|}{}                       & \multicolumn{1}{c|}{}                         & \multicolumn{3}{c|}{0}                                                                                             & \multicolumn{4}{c|}{500}                                                                                           & \multicolumn{4}{c|}{1000}                                                                                          & \multicolumn{4}{c}{2000}                                                                                      \\ \cline{3-17} 
\multicolumn{1}{c|}{\multirow{-3}{*}{Data}} & \multicolumn{1}{c|}{\multirow{-3}{*}{Method}} & \multicolumn{1}{l}{PSNR}      & \multicolumn{1}{l}{SSIM}      & \multicolumn{1}{l|}{MSIM}                          & \multicolumn{1}{l}{PSNR}      & \multicolumn{1}{l}{SSIM}      & \multicolumn{1}{l}{MSIM}   & \multicolumn{1}{l|}{SIZE}                       & \multicolumn{1}{l}{PSNR}      & \multicolumn{1}{l}{SSIM}      & \multicolumn{1}{l}{MSIM} & \multicolumn{1}{l|}{SIZE}                         & \multicolumn{1}{l}{PSNR}      & \multicolumn{1}{l}{SSIM}      & \multicolumn{1}{l}{MSIM} & \multicolumn{1}{l}{SIZE}                           \\ \hline
                                            & Triplanes                                     & 11.36                          & 0.784                         & \multicolumn{1}{c|}{0.330}                         & 22.23                         & 0.863           & 0.887              & \multicolumn{1}{c|}{8.077}                         & 25.27                         & 0.904                   & 0.957      & \multicolumn{1}{c|}{8.077}                         & 26.56                         & 0.921                & 0.959         & \multicolumn{1}{c}{8.077}                           \\
                                            & \textbf{Ours (PEFT)}                     & \cellcolor[HTML]{FFCCC9}22.45 & \cellcolor[HTML]{FFCCC9}0.871 & \multicolumn{1}{c|}{\cellcolor[HTML]{FFCCC9}0.901} & \cellcolor[HTML]{FFCCC9}25.95 & \cellcolor[HTML]{FFCCC9}0.902 & \cellcolor[HTML]{FFCCC9}0.945 & \multicolumn{1}{c|}{1.024} & \cellcolor[HTML]{FFCCC9}27.61 & \cellcolor[HTML]{FFCCC9}0.921 & \cellcolor[HTML]{FFCCC9}0.961 & \multicolumn{1}{c|}{1.024} & \cellcolor[HTML]{FFCCC9}28.57 & \cellcolor[HTML]{FFCCC9}0.932 & \cellcolor[HTML]{FFCCC9}0.968 & \multicolumn{1}{l}{1.024}\\
\multirow{-3}{*}{Obj}                     & \textbf{Ours (PEFT++)}                   & $\cdot$                         & $\cdot$                         & \multicolumn{1}{c|}{$\cdot$}                         & 25.79                         & 0.901                         & 0.943 & \multicolumn{1}{c|}{\cellcolor[HTML]{FFCCC9}0.197}                         & 27.35                         & 0.918                         & 0.959 & \multicolumn{1}{c|}{\cellcolor[HTML]{FFCCC9}0.179} & 28.28                         & 0.929 & 0.967 & \multicolumn{1}{l}{\cellcolor[HTML]{FFCCC9}0.146}  \\ \hline
                                            & Triplanes                                     & 12.55                         & 0.836                         & \multicolumn{1}{c|}{0.364}                         & 28.32                         & 0.940                         & 0.964  & \multicolumn{1}{c|}{8.077}                          & 30.13                         & 0.955                         & 0.970  & \multicolumn{1}{c|}{8.077}                         & 31.54                         & 0.964                         & 0.975 & \multicolumn{1}{l}{8.077}                            \\
                                            & \textbf{Ours (PEFT)}                     & \cellcolor[HTML]{FFCCC9}23.98 & \cellcolor[HTML]{FFCCC9}0.892 & \multicolumn{1}{c|}{\cellcolor[HTML]{FFCCC9}0.914} & \cellcolor[HTML]{FFCCC9}31.90 & \cellcolor[HTML]{FFCCC9}0.952 & \cellcolor[HTML]{FFCCC9}0.981 & \multicolumn{1}{c|}{1.024} & \cellcolor[HTML]{FFCCC9}33.35 & \cellcolor[HTML]{FFCCC9}0.961 & \cellcolor[HTML]{FFCCC9}0.986 & \multicolumn{1}{c|}{1.024} & \cellcolor[HTML]{FFCCC9}34.65 & \cellcolor[HTML]{FFCCC9}0.968 & \cellcolor[HTML]{FFCCC9}0.990 & \multicolumn{1}{l}{1.024}      \\
\multirow{-3}{*}{GSO}                       & \textbf{Ours (PEFT++)}                   & $\cdot$                         & $\cdot$                         & \multicolumn{1}{c|}{$\cdot$}                         & 31.38                         & 0.949                         & 0.978 & \multicolumn{1}{c|}{\cellcolor[HTML]{FFCCC9}0.188}                         & 32.70                         & 0.958                         & 0.984 & \multicolumn{1}{c|}{\cellcolor[HTML]{FFCCC9}0.172}                         & 33.87                         & 0.965                         & 0.988  & \multicolumn{1}{l}{\cellcolor[HTML]{FFCCC9}0.145}
\end{tabular}%
}
\caption{Quantitative results of the proposed methods evaluated on Objaverse (denoted by `Obj') and GSO datasets. `PEFT++' denotes parameter efficient finetuning with entropy coding, `MSIM' is MSSSIM and `SIZE' is measured in MB.}
\label{table:quan}
\end{table*}

\subsection{Training objective}
\label{sec:training_loss}
Here, we train our base model in an end-to-end manner with its fully differentiable properties. We use L2 loss, denoted as $\mathcal{L}_{\mathrm{rgb}}$ to measure the pixel-level difference and LPIPS~\cite{zhang2018unreasonable} loss, denoted as $\mathcal{L}_{\mathrm{lpips}}$ to measure the patch-level difference between the ground truth images and rendered images.


Spatial total variation (TV) regularization encourages the sparse or smooth gradient, thereby ensuring that the feature planes do not contain erroneous high-frequency data~\cite{kplanes_2023, chen2022tensorf, shue20223nfd}. We use the standard L2 TV regularization as default to make the distribution of the triplane features smoother, as it regularizes the squared difference between the neighboring values in the feature maps.
\begin{align}
    \mathcal{L}_{\mathrm{tv}}=\frac{1}{T} \hspace{-0.5em}\sum_{k \in \{xy,yz,xz\}} \sum_{s=1}^2 \sum_{i,j} &\left(\left\|\tilde{f}_{k,i,j}^s-\tilde{f}_{k,i-1,j}^s\right\|_2^2 \right. \notag \\
    &\left. + \left\|\tilde{f}_{k,i,j}^s-\tilde{f}_{k,i,j-1}^s\right\|_2^2 \right)    
\end{align}
where $T=3C(V_1^2 + V_2^2)$ is the total number of features across all triplanes and resolutions. The final objective function for a training instance can be written as follows,
\begin{equation}
\mathcal{L}=\mathcal{L}_{\mathrm{rgb}}+\mathcal{L}_{\mathrm{vq}}+\lambda_\mathrm{lpips} \mathcal{L}_{\mathrm{lpips}}+\lambda_\mathrm{tv} \mathcal{L}_{\mathrm{tv}}.
\end{equation}



\subsection{Parameter-efficient finetuning}
\label{sec:peft}
While the NeRF representations generated by the encoder and decoder modules are of high quality, their generalization performance on a new scene can be limited.
Similar to other NeRF generalization models~\cite{tancik2021learnedinit, bergman2021fastlumigraph}, our approach can also leverage the finetuning of NeRF representations to enhance visual quality for new scenes during testing time.
However, effective model finetuning is severely hindered by the growing computational costs and memory storage as the model size increases.
To tackle this issue, LoRA~\cite{hu2021lora} is a widely used parameter-efficient finetuning (PEFT) method for adaptation of large-scale models, mainly explored in NLP and computer vision domains.
We propose to adapt PEFT in our test time optimization, and to the best of our knowledge, we are the first to apply PEFT to 3D NeRF representation.
We first generate multi-resolution triplanes using multi-view test images, and train only the triplanes and decoder in an efficient way.

\noindent{\bf Parameter efficient triplane finetuning.}
We propose a tensor factorization scheme to efficiently finetune triplane representation. Let $\tilde{f}_{k}^{s} \in \mathbb{R}^{C \times V_s \times V_s}$, be the triplanes generated by the encoder and decoder for a scale `$s$' and plane `$k$' ($s \in \{ 1, 2\}$ and $k \in \{xy,yz,xz\}$). The final triplane representations are expressed by $\tilde{f}_k^{s} + \Delta \tilde{f}_k^{s}$, and $\Delta \tilde{f}_k^{s} \in \mathbb{R}^{C \times V_s \times V_s}$ is constructed by a tensor product between matrices and vectors.
\begin{equation}
\Delta \tilde{f}_{k}^{s}=\sum_{r=1}^R v_r^s \circ M_{k,r}^{s},
\end{equation}

where $M_{k,r}^{s} \in \mathbb{R}^{V_s \times V_s}$ denotes $r$-th matrix for the `$k$' plane and scale `$s$',  $v_r^s \in \mathbb{R}^{C}$ is the $r$-th vector for all three planes and scale `$s$', and $\circ:\mathbb{R}^{C} \times \mathbb{R}^{V_s \times V_s} \rightarrow \mathbb{R}^{C \times V_s \times V_s}$ is a tensor product. During finetuning, we freeze $\tilde{f}_{k}^{s}$ and only updates $\Delta \tilde{f}_{k}^{s}$. We apply this scheme for every feature planes in multi-resolution triplanes and used $R=4$ (for ablation studies over different rank size, please refer to the supplementary materials). For initialization, we use a common technique, setting all matrices to random values and all vectors to zeros. It makes our delta to be zero at the start of the training.

\noindent{\bf Parameter efficient MLP finetuning.}
Additionally, we factorize MLP layers in decoders using the LoRA method for MLP finetuning. Using two PEFT methods, we can achieve massive reductions in trainable parameters during test time optimization (Fig.~\ref{fig:PEFT}). The training objective is the same with the base model except for the vector quantization and LPIPS losses.

\subsection{Entropy coding finetuning deltas}
\label{sec:peftec}
We leverage neural compression methods that have demonstrated efficacy in image and video domains to seek to achieve the optimal compression rate~\cite{balle2016compiclr17, balle2018compiclr18, li2021dcvc, lin2020mlvc, sheng2022temporal}. We adopt a entropy coding model~\cite{balle2018compiclr18} to our proposed parameter-efficient finetuning of the triplanes. We model the prior using a non-parametric density, which is convolved with a standard uniform density in a differentiable manner (please refer to \cite{balle2018compiclr18} for more details). 
Then, our training objective for the finetuning is defined as follows.
\begin{align}
&\mathcal{L}_{\mathrm{ent}} = \sum_{k \in \{xy,yz,xz\}} \sum_{r=1}^R \sum_{s=1}^2 -\log p(M_{k,r}^{s}), \\
&\mathcal{L}=\mathcal{L}_{\mathrm{rgb}}+\lambda_\mathrm{rate} \mathcal{L}_{\mathrm{ent}}+\lambda_\mathrm{tv} \mathcal{L}_{\mathrm{tv}},
\end{align}
where $\lambda_\mathrm{rate}$ will balance between the quantization error and the code length.
We only applied the entropy model to the feature matrices for triplane features, and after finetuning, updated matrices $M_{k,r}^{s}$ are quantized and compressed, while other weights, $\omega_a, \omega_b, v_r^s$ are stored with 32-bit precision as a convention. 
Please see the supplementary for the details and entropy coding on MLP (LoRA).


\begin{figure*}[!ht]
  \centering
  \includegraphics[width=0.95\linewidth]{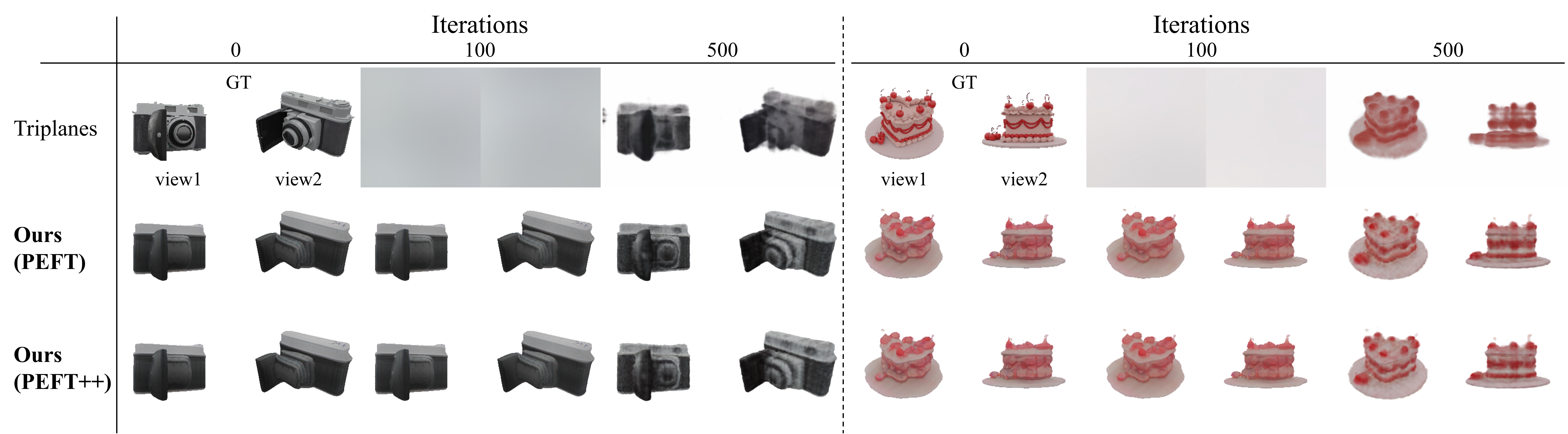}
  \caption{Novel view synthesis results on Objaverse dataset.}
  \label{fig:results}
\end{figure*}

\section{Experiments}
\subsection{Datasets}
To evaluate our method, we conduct experiments on 1) Objaverse~\cite{deitke2023objaverse} and Google Scanned Objects (GSO)~\cite{downs2022google} for object-level novel view synthesis, and 2) DTU dataset~\cite{jensen2014dtu} for real scenes. For Objaverse, we sourced images from One-2-3-45~\cite{Liu202312345}, which consists of 46k objects, and constructed our own split of 36,796 training objects and 9,118 test objects. In GSO, we used 1,030 objects only for the evaluation. Lastly, we followed PixelNeRF~\cite{yu2021pixelnerf} DTU dataset split with 88 training scenes and 15 testing scenes. Object images are rendered at a resolution of 256$\times$256 and we cropped the DTU images to match the same resolution.



\subsection{Implementation Details}
\label{imple}
To train our base model, we randomly choose 16 input images and camera poses to produce a triplane representation and predict the remaining novel views. 
We used two spatial resolutions \{$V_{1}, V_{2}$\} = \{64, 128\} and channel size $C$ = 32 for our multi-resolution triplanes. The MLP decoders are of 6 layers with hidden dimensions 64 for coarse and fine decoders, respectively. We set the codebook size $K=8192$ and dimension $C'=32$.
In the finetuning stage, we first generated an initial representation using predetermined 16 view indices and finetuned the triplane features with MLP decoders in an optimization-based approach. We finetuned the model using 24 images and tested it on the remaining views. Note that the 16 images used to generate the initial triplane representations are a subset of the 24 training images, and the same images are all used to train baselines for a fair comparison. For the DTU dataset, we choose 8 input images for base model training and 16 images for finetuning. We set the LoRA's rank to 4 in every layer of the decoder. 

\subsection{Results}

\label{sec:results}
\noindent{\bf Object-level Benchmarks.} To assess the efficacy of our method in test time optimization, we employed K-planes~\cite{kplanes_2023} as our baseline model, which has shown fast training and compact representation. We revised the architecture based on our method for a fair comparison, and this model will be referred to as `Triplanes'.
We evaluated the performance under two different scenarios starting from our generated triplane initializations: 1) parameter-efficient finetuning (PEFT) and 2) parameter-efficient finetuning with the proposed entropy coding (PEFT++). For further results and detailed configurations, consult the supplementary materials.

For the quantitative metrics, we report the standard image quality measurements, PSNR, SSIM, and MS-SSIM. 
We also measure the storage requirements of the representations to show the compression performance of our method. 
As shown in Tab.~\ref{table:quan}, our method shows fast encoding progress from its initialized representation and improvement on all metrics. Thanks to the pretrained generalization capability of our method, our model outperforms the per-scene optimization baseline, Triplanes, in novel view synthesis. 
The qualitative results in Fig.~\ref{fig:results} present the novel view synthesis results across the finetuning iteration, illustrating the generalization ability and fast encoding speed of our methods. In Tab.~\ref{table:compression}, we report the component-level storage comparison (on the Objaverse dataset) after 10k iterations. Our method achieved 100$\times$ compactness with better quality compared to the baseline model.

\begin{figure}[!t]
  \centering
  \includegraphics[width=0.9\linewidth]{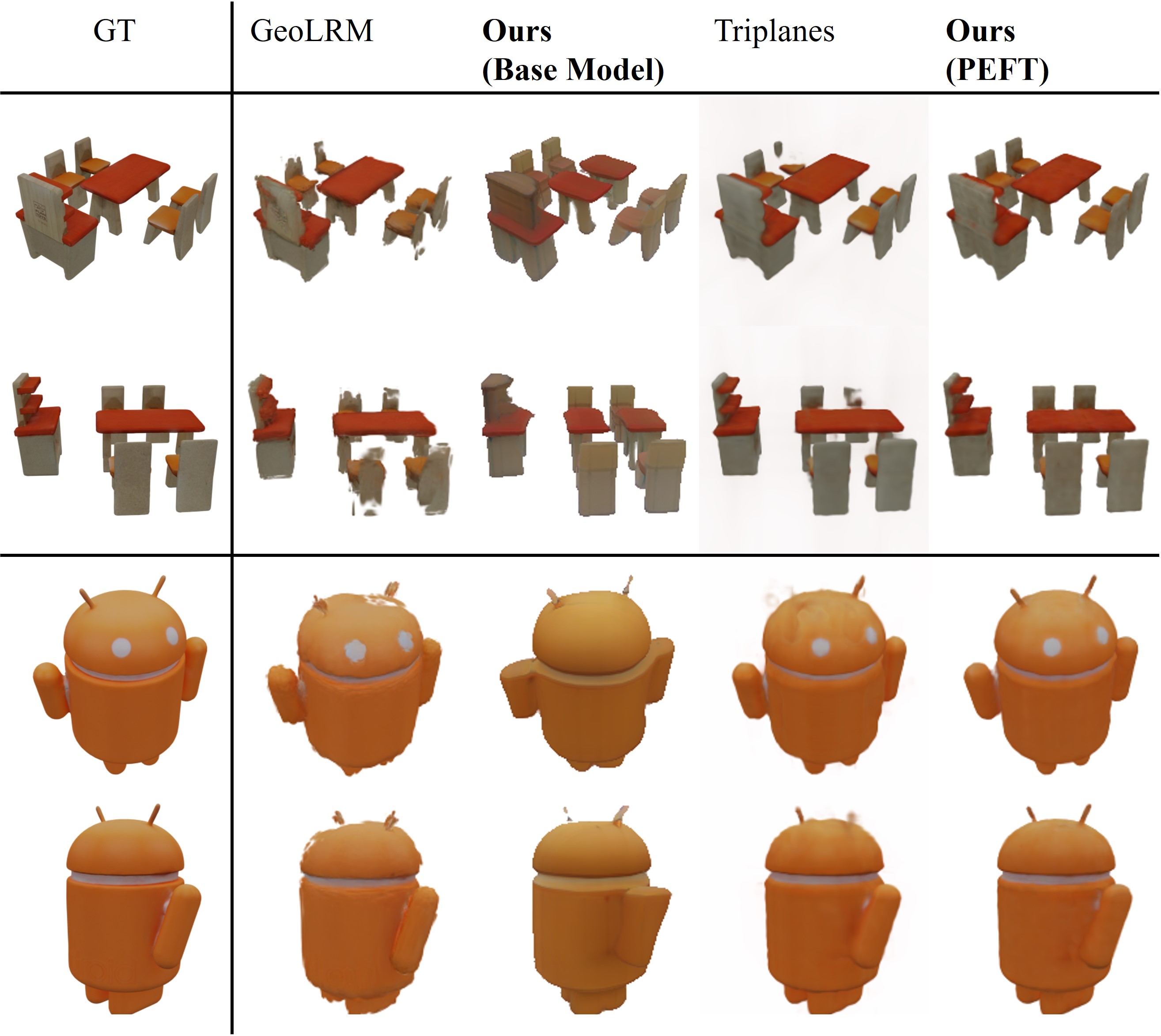}
  \caption{Novel view synthesis results on GSO dataset. `Triplanes' and `Ours (PEFT)' is finetuned with 1k iterations.}
  \label{fig:geo_test}
\end{figure}

We also compared our model with a large reconstruction model, GeoLRM~\cite{zhang2024geolrm}, a Gaussian-based reconstruction model. Since GeoLRM has the property that can scale up to dense views, we choose it as the baseline for a fair comparison with our model. We utilized the GSO dataset, and the same 16 views were used as input, and the remaining views were used to evaluate the performance. Tab.~\ref{table:table_gso} shows that our model outperforms the baseline on PSNR and SSIM while often missing the detailed visual quality as depicted in Fig.~\ref{fig:geo_test}. We suspect that the small code size in the bottleneck layer incurs the difficulty in providing enough details. Nevertheless, after finetuning with our proposed method, we can attain high-quality 3D representations quickly with a very compact size, as shown in Tab.~\ref{table:quan} and Fig.~\ref{fig:geo_test}. We evaluated all objects in Tab.~\ref{table:table_gso}, and 100 objects in Tab.~\ref{table:quan}


\begin{table}[h!]
\centering
\renewcommand{\arraystretch}{1.0}
\resizebox{0.9\linewidth}{!}{
\begin{tabular}{c|cccc}
\multirow{2}{*}{Component} & \multicolumn{4}{c}{Total size in MB (codes + finetuning deltas)}                                                                                                             \\ \cline{2-5} 
                           & \multicolumn{1}{c|}{Triplanes} &  \multicolumn{1}{c|}{PEFT} & \multicolumn{1}{c|}{PEFT++} & \multicolumn{1}{l}{W/O FT} \\ \hline
Codes                   & \multicolumn{1}{c|}{.}        & \multicolumn{1}{c|}{0.013}         & \multicolumn{1}{c|}{0.013}      & 0.013                         \\
Feature                   & \multicolumn{1}{c|}{7.864}          & \multicolumn{1}{c|}{\cellcolor[HTML]{FFFC9E}0.984}   & \multicolumn{1}{c|}{\cellcolor[HTML]{FFCCC9}0.031}     & .                         \\
MLP                        & \multicolumn{1}{c|}{0.213}           & \multicolumn{1}{c|}{\cellcolor[HTML]{FFCCC9}0.027}   & \multicolumn{1}{c|}{\cellcolor[HTML]{FFCCC9}0.027}     & .                         \\ \hline
Total   & \multicolumn{1}{c|}{8.077}          & \multicolumn{1}{c|}{\cellcolor[HTML]{FFFC9E}1.024}    & \multicolumn{1}{c|}{\cellcolor[HTML]{FFCCC9}0.071}     & 0.013 
\end{tabular}}
\caption{Memory footprint in component level.}
\label{table:compression}
\end{table}

\begin{table}[h!]
\centering
\resizebox{0.85\linewidth}{!}{
\Large
\begin{tabular}{c|cccc}
\multicolumn{1}{l|}{} & \multicolumn{1}{l}{PSNR} & \multicolumn{1}{l}{SSIM} & \multicolumn{1}{l}{MSSSIM} & \multicolumn{1}{l}{Codes (MB)} \\ \hline
GeoLRM            & 23.73                        & 0.890                                          & \cellcolor[HTML]{FFCCC9}0.922   & .      \\
Ours                   & \cellcolor[HTML]{FFCCC9}24.49  & \cellcolor[HTML]{FFCCC9}0.897                                                              & 0.918  & 0.015                   \\
\end{tabular}%
}
\caption{Initialization performance on GSO dataset.}
\label{table:table_gso}
\end{table}


\begin{figure}[h!]
  \centering
  \includegraphics[width=\linewidth]{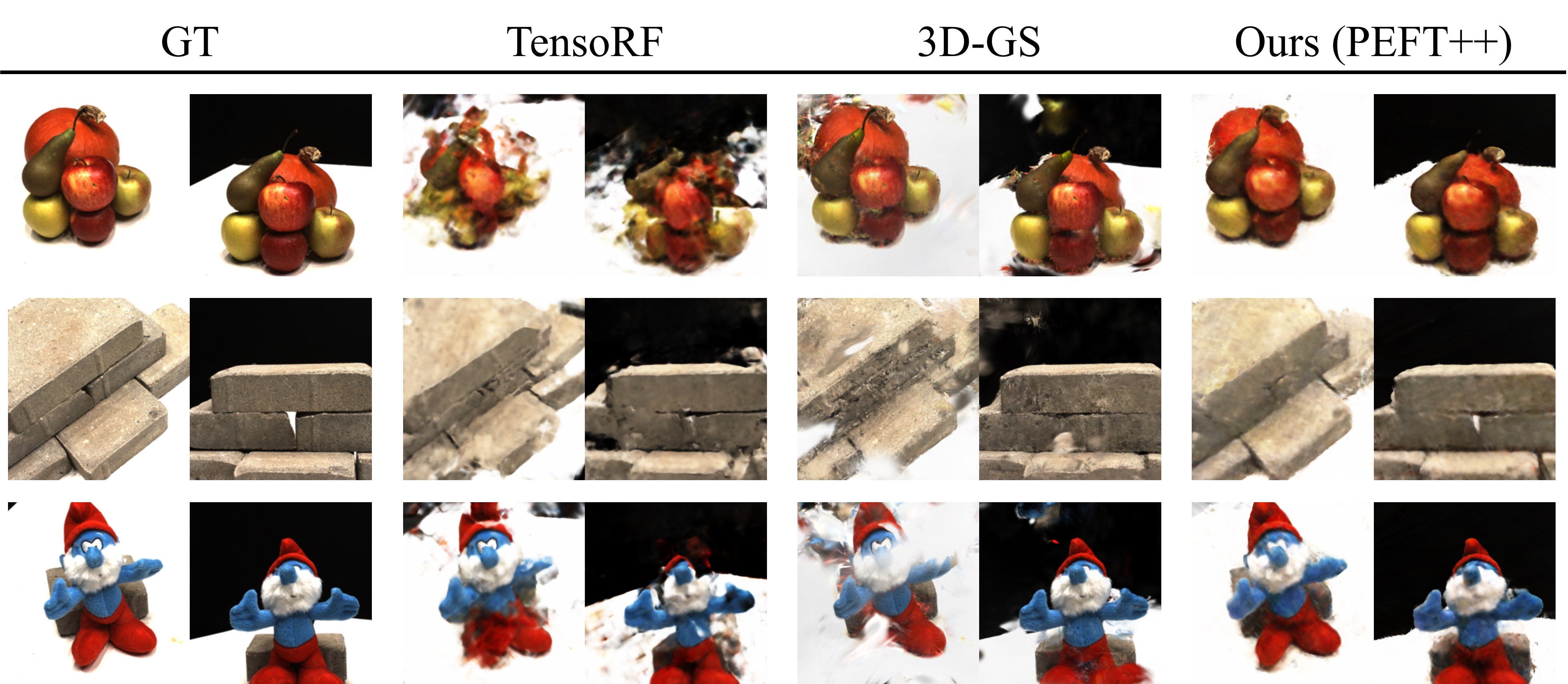}
  \caption{Novel view synthesis results on DTU dataset.}
  \label{fig:dtu_results}
\end{figure}

\noindent{\bf Scene-level Benchmarks.} We further demonstrate the applicability of our method on real scenes using the DTU dataset. Two representative optimization-based methods, TensoRF~\cite{chen2022tensorf} and 3D-GS~\cite{kerbl20233dgs}, are used as baselines. As shown in Tab.~\ref{table:table_dtu}, our method was finetuned for less than a minute to evaluate its fast encoding ability. Fig.~\ref{fig:dtu_results} presents the novel view synthesis results on the DTU dataset. Both quantitative and qualitative results indicate that our method surpasses TensoRF and 3D-GS in terms of quality, even with a compact representation size. 

\begin{table}[h!]
\centering
\renewcommand{\arraystretch}{1.0}
\footnotesize{
\resizebox{0.9\columnwidth}{!}{
\begin{tabular}{l|cccc}
Method                          & \multicolumn{1}{l}{PSNR} & \multicolumn{1}{l}{SSIM} & \multicolumn{1}{l}{Train (s)} & \multicolumn{1}{l}{Size (MB)} \\ \hline
\multirow{1}{*}{TensoRF}       & 14.58                    & 0.517                    & 57.3                         & 5.430                         \\
\multirow{1}{*}{3D-GS}         & 15.77                    & 0.581                     & 64.0                          & 156.9                         \\ \hline
\multirow{1}{*}{\textbf{Ours (PEFT)}}   & 20.05                    & \cellcolor[HTML]{FFCCC9}0.650                    & \cellcolor[HTML]{FFCCC9}41.3                          & 1.023                          \\
\multirow{1}{*}{\textbf{Ours (PEFT++)}} & \cellcolor[HTML]{FFCCC9}20.07                    & 0.646                    & 58.8                          & \cellcolor[HTML]{FFCCC9}0.160                       \\               
\end{tabular}}}
\caption{Perfromance comparison on DTU dataset.}
\label{table:table_dtu}
\end{table}
\vspace{-5mm}

\begin{figure}[h]
  \centering
  \includegraphics[width=85mm]{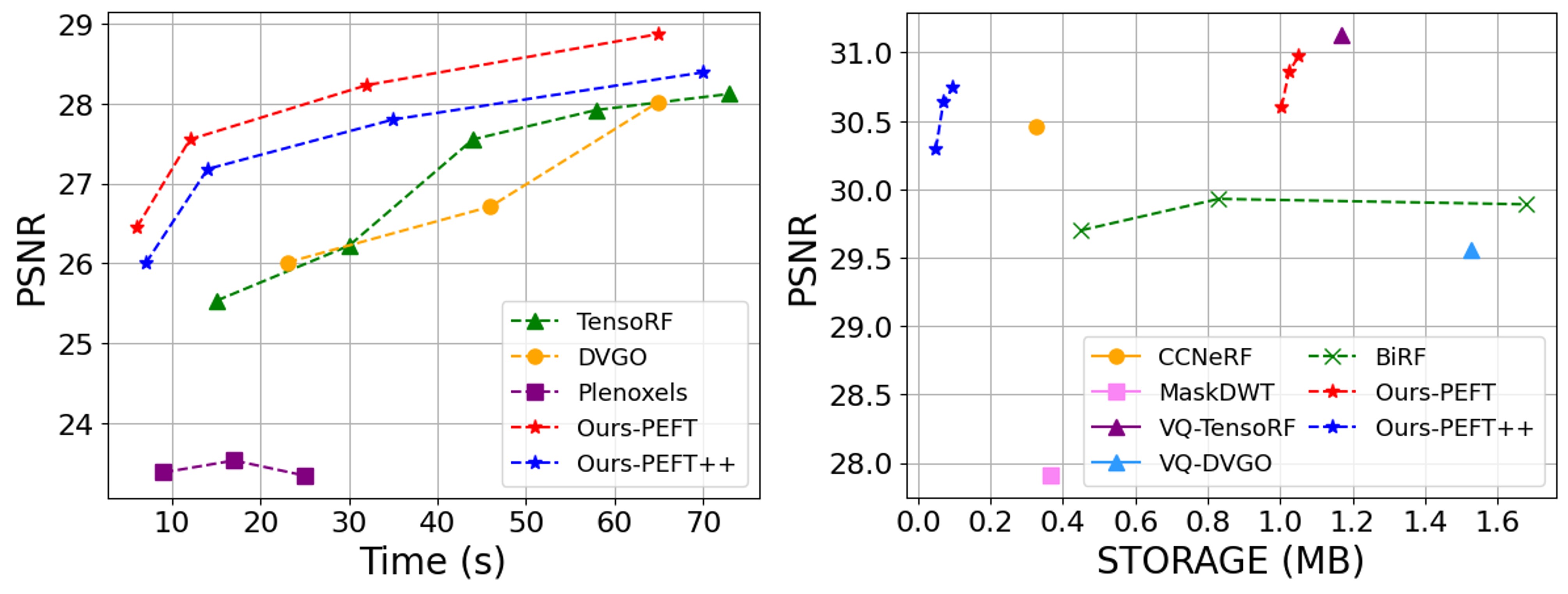}
  \caption{In-depth evaluations.}
  \label{fig:in-depth}
\end{figure}

\noindent{\bf In-depth Evaluations.} We conducted a detailed comparison of our method with both fast and compact specialized models on Objaverse dataset. For fast training NeRF, we adopted TensoRF~\cite{chen2022tensorf}, DVGO~\cite{sun2022dvgo}, and Plenoxels~\cite{fridovich2022plenoxels} and for compact NeRF, we employed CCNeRF~\cite{tang2022compressible}, MaskDWT~\cite{rho2023maskedwavelet}, VQRF~\cite{li2023compressing1mb}, and BiRF~\cite{shin2024binarynerf}. We showed different compression rates with our method by varing the rank size in decoder's LoRA (1, 4, and 8).  As illustrated in Fig.~\ref{fig:in-depth}, our method demonstrates the ability to achieve fast encoding and a high compression ratio, outperforming representative models in both aspects.

\begin{figure}[hbt!]
  \centering
  \includegraphics[width=1.0\linewidth]{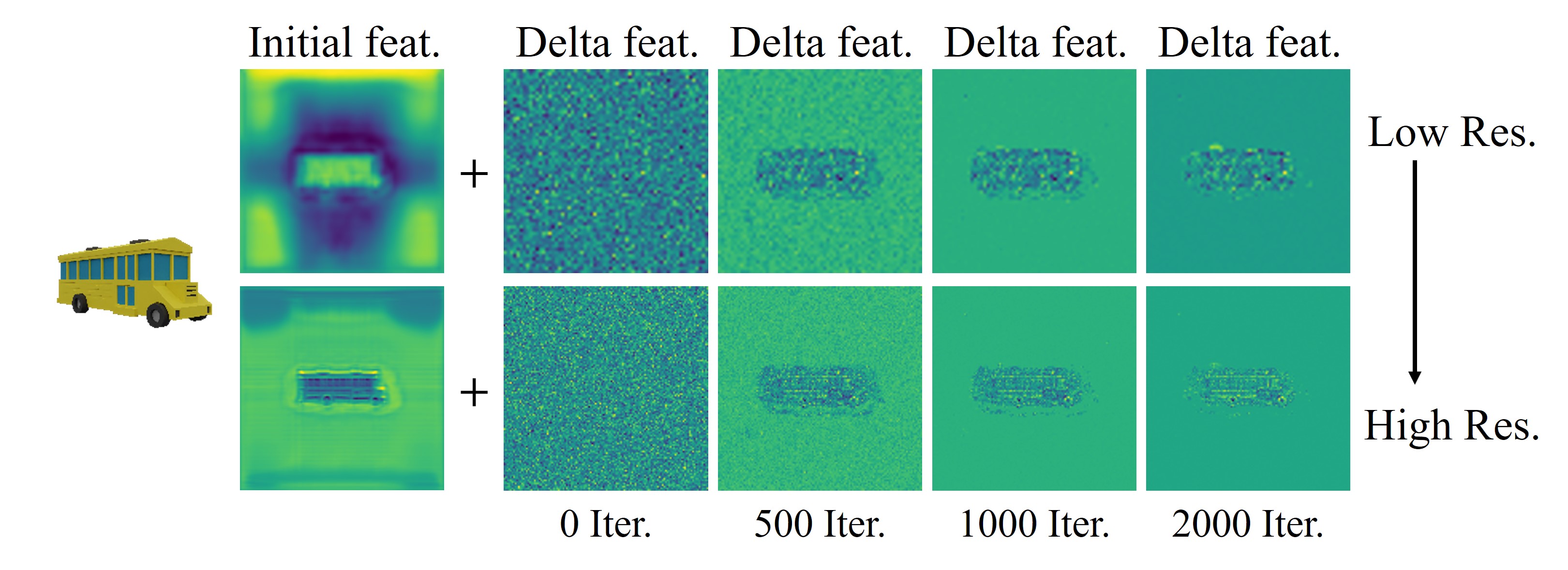}
  \caption{Visualization of delta feature maps finetuned with entropy coding. The averaged $YZ$ planes across the channel dimensions are shown in the different resolutions.}
  \label{fig:featuremap}
\end{figure}

\noindent{\bf{Feature Visualization}} We also visualized the delta feature maps across finetuning iterations based on our entropy coding method. As shown in Fig.~\ref{fig:featuremap}, the feature maps following the entropy coding, eliminate unnecessary components at different resolutions, and get a high compression ratio resulting from quantization. This observation shows that employing different spatial resolutions would help reduce the quantity of information stored at each level, thus making the use of the entropy coding as an ideal strategy.

\section{Conclusion}
\label{sec:conclusion}
In this work, we introduced CodecNeRF, a novel encoding-decoding-finetuning pipeline designed for fast encoding and decoding, compact codes, and high-quality renderings.
To our knowledge, this is the first attempt to propose a neural learned codec for emerging 3D representations, such as NeRF.
Our experimental results demonstrated a significant performance improvement over strong NeRF baseline models across commonly used 3D object datasets, including Objaverse and Google Scanned Objects, as well as the scene-level DTU dataset. We believe that our framework has the potential to pave the way for new research directions and broaden the applications of NeRF.

\clearpage
\appendix
\begin{center}
    \Large \textbf{Supplementary Materials for CodecNeRF}
\end{center}
\section{Ablation Studies}
\label{sec:ablation_studies}
This section describes three different ablation scenarios comparing with our default method: 1) different feature decomposition, 2) applying entropy coding on MLP decoder, and 3) component for finetuning.
\subsection{Feature decomposition}
\label{sub:feature_decomposition}
\begin{figure}[!h]
\begin{minipage}[c]{.69\linewidth}
\centering
\includegraphics[width =0.9\textwidth]{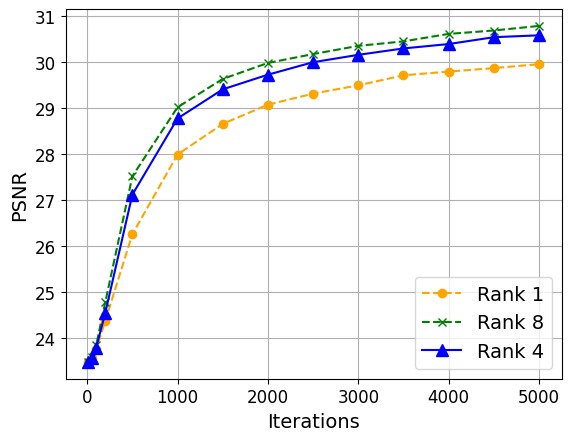}
\caption{Ablation over decomposition.}
\label{fig:albation_decomp}
\end{minipage}
\begin{minipage}[c]{.29\linewidth}
\centering
\captionsetup{type=table} 
\vspace{4mm}
\renewcommand{\arraystretch}{1.1}
\resizebox{\textwidth}{!}{%
\begin{tabular}{|c|c|}
\hline
Method        & Storage (MB) \\ \hline
Rank1 & 0.007        \\ \hline
\textbf{Rank 4}             & 0.027        \\ \hline
Rank 8             & 0.056        \\ \hline
\end{tabular}}
\caption{Storage.}
\label{tab:decomp_table}
\end{minipage}
\end{figure}

In Tab.~\ref{tab:decomp_table},~\ref{tab:ablation_decomp} and Fig.~\ref{fig:albation_decomp}, we ablated our method on the Objaverse dataset~\cite{deitke2023objaverse}, with respect to varying rank size in tensor decomposition. Our PEFT method is used in experiments: all training settings are fixed except for the rank size. We selected our configuration considering the trade-offs between encoding speed and compression ratio.

\begin{table}[!h]
\centering
\renewcommand{\arraystretch}{1.2}
\resizebox{\columnwidth}{!}{
\begin{tabular}{l|ccccccccc}
\multicolumn{1}{c|}{\multirow{3}{*}{Method}} & \multicolumn{9}{c}{Iterations}                                                                                \\ \cline{2-10} 
\multicolumn{1}{c|}{}                        & \multicolumn{3}{c|}{500}                & \multicolumn{3}{c|}{1000}               & \multicolumn{3}{c}{2000} \\ \cline{2-10} 
\multicolumn{1}{c|}{}                        & PSNR & SSIM & \multicolumn{1}{c|}{MSIM} & PSNR & SSIM & \multicolumn{1}{c|}{MSIM} & PSNR   & SSIM   & MSIM   \\ \hline
Rank 1                                       &  26.25    & 0.896     & \multicolumn{1}{c|}{0.943}     & 27.99     & 0.915     & \multicolumn{1}{c|}{0.960}     & 29.07       & 0.927       & 0.969      \\
\textbf{Rank 4}                              & 27.11     & 0.905     & \multicolumn{1}{c|}{0.953}     & 28.77     & 0.923     & \multicolumn{1}{c|}{0.967}     & 29.72       & 0.933       & 0.974       \\
Rank 8                                       & 27.51     & 0.909     & \multicolumn{1}{c|}{0.957}     & 29.01     & 0.926     & \multicolumn{1}{c|}{0.970}     & 29.98       & 0.936       & 0.976  
\end{tabular}}
\caption{Quantitative results on different decomposition rank on objaverse dataset, `MISM' is MSSSIM.}
\label{tab:ablation_decomp}
\end{table}

\subsection{Entropy coding on decoder}
\label{sub:mlp_entropy_coding}
\begin{figure}
  \centering
  \includegraphics[width=\linewidth]{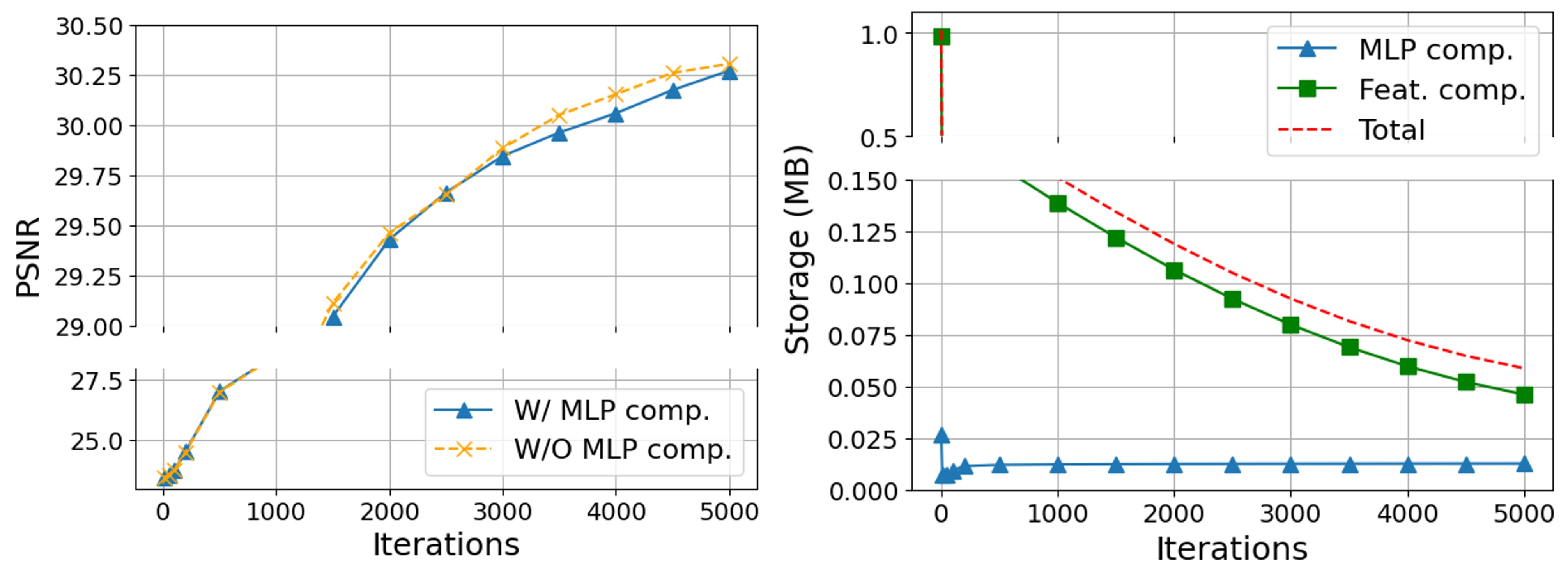}
  \caption{Ablation over weight entropy coding.}
  \label{fig:ablation_weight}
\end{figure}

\begin{figure*}[!h]
  \centering
  \includegraphics[width=1.0\linewidth]{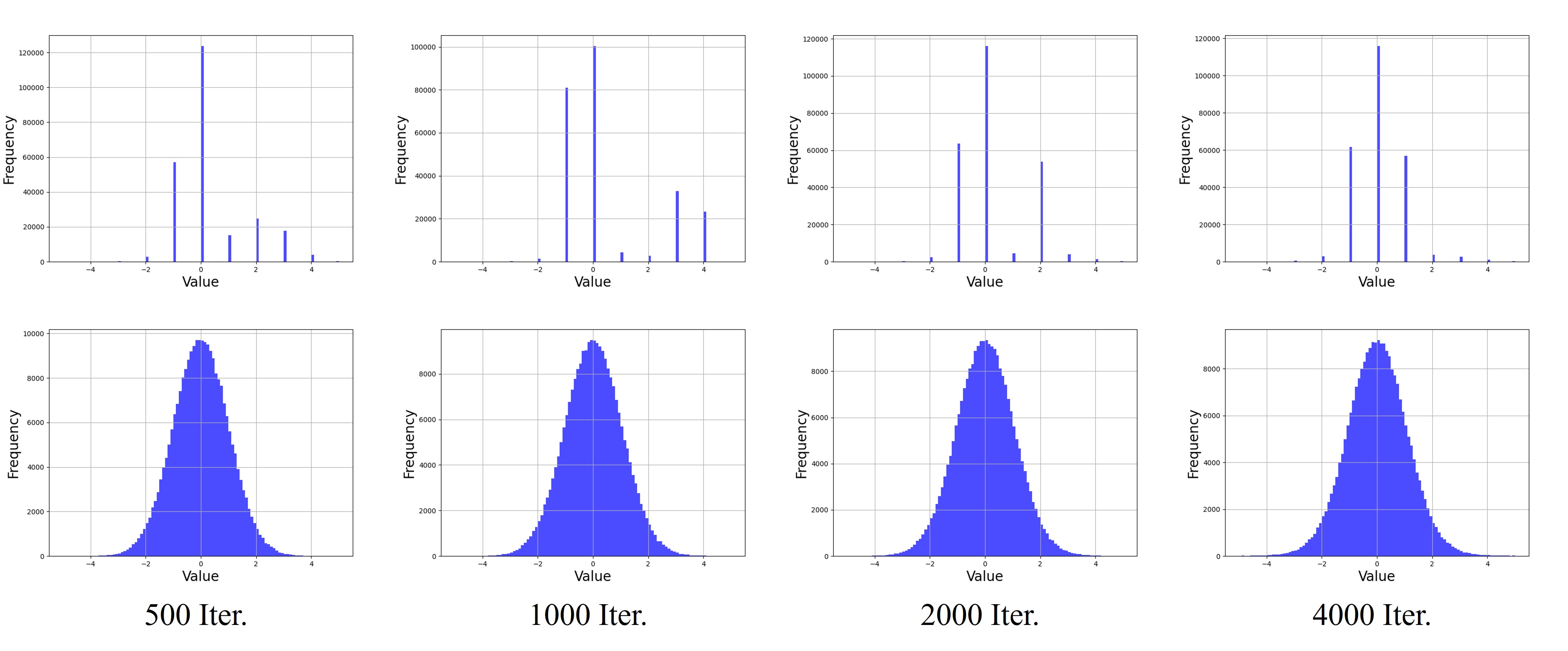}
  \caption{Feature value histogram. We used decomposed feature matrix for the visualization. The first row dispicts the histogram with entropy coding, and the next row shows the histogram without entropy coding.}
  \label{fig:ablation_tensor_hist}
\end{figure*}

\begin{figure*}[!h]
  \centering
  \includegraphics[width=1.0\linewidth]{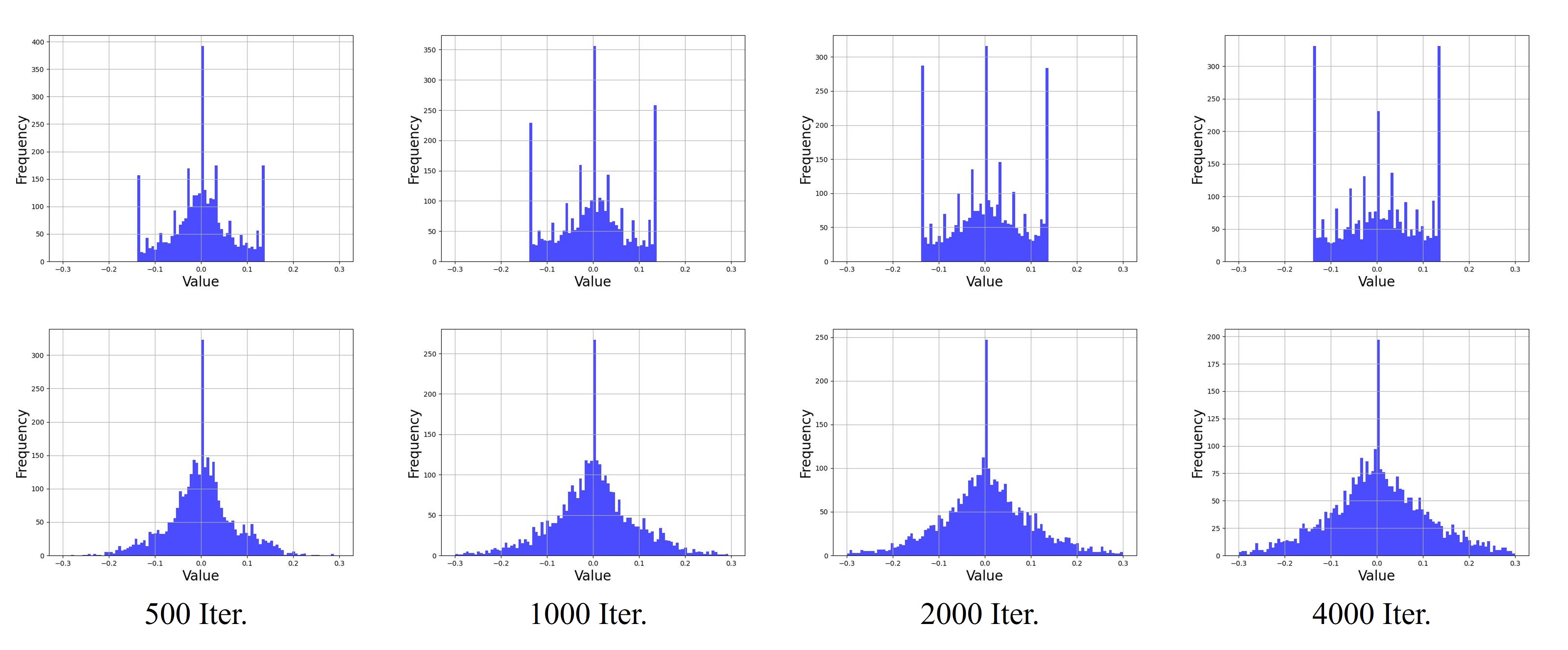}
  \caption{Decoder value histogram. We visualized the histogram with the fine decoder weights of LoRA. The first row dispicts the results with weight entropy model, and the next row shows the histogram without weight entropy model.}
  \label{fig:ablation_weight_hist}
\end{figure*}

We ablated our method on the Objaverse dataeset, with respect to the entropy coding on MLP decoder, LoRA~\cite{hu2021lora}. We used a spike-and-slab prior~\cite{rovckova2018spike}, a mixture of two Gaussians (a wide and a narrow distributions), to approximate the entropy and compress the decoder weights. Our PEFT++ method is used in experiments. As shown in Fig.~\ref{fig:ablation_weight} and Tab.~\ref{tab:ablation_weight}, the degradation in performance is negligible, while the final code length of the LoRA weights is decreased. Although it increases along the iterations after the first drop, it obviously requires less storage than the unapplied version. We also visualized the histogram of the decomposed triplane features and LoRA weights after the entropy coding. Fig.~\ref{fig:ablation_tensor_hist} and~\ref{fig:ablation_weight_hist} describe the progress of compression across the iterations. 

\begin{table}
\centering
\renewcommand{\arraystretch}{1.2}
\resizebox{\columnwidth}{!}{
\begin{tabular}{l|ccccccccc}
\multicolumn{1}{c|}{\multirow{3}{*}{Method}} & \multicolumn{9}{c}{Iterations}                                                                                \\ \cline{2-10} 
\multicolumn{1}{c|}{}                        & \multicolumn{3}{c|}{500}                & \multicolumn{3}{c|}{1000}               & \multicolumn{3}{c}{2000} \\ \cline{2-10} 
\multicolumn{1}{c|}{}                        & PSNR & SSIM & \multicolumn{1}{c|}{MSIM} & PSNR & SSIM & \multicolumn{1}{c|}{MSIM} & PSNR   & SSIM   & MSIM   \\ \hline
\textbf{W/ entropy}                                       &  27.04    & 0.905     & \multicolumn{1}{c|}{0.952}     & 28.54     & 0.921     & \multicolumn{1}{c|}{0.965}     & 29.43       & 0.930       & 0.972      \\
W/O entropy                              & 27.00     & 0.904     & \multicolumn{1}{c|}{0.951}     & 28.53     & 0.921     & \multicolumn{1}{c|}{0.965}     & 29.46       & 0.931       & 0.973  
\end{tabular}}
\caption{Quantitative results on entropy coding upon decoder weight (LoRA). `MISM' is MSSSIM.}
\label{tab:ablation_weight}
\end{table}

\subsection{Finetuning component}
\label{sub:finetuning_component}
\begin{figure}[!hbt]
  \centering
  \includegraphics[width=0.8\linewidth]{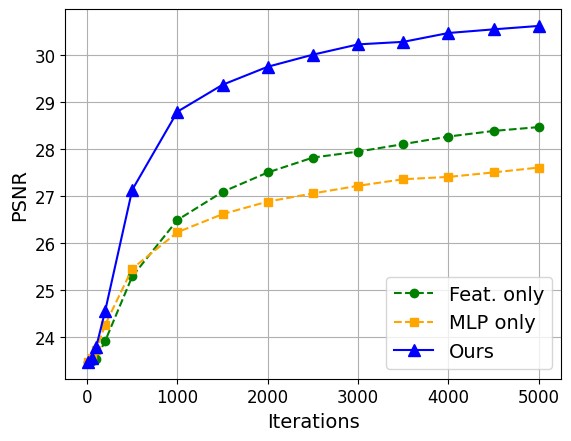}
  \caption{Ablation over finetuning component.}
  \label{fig:ablation_component}
\end{figure}

We ablated our finetuning method on the objaverse dataset, with respect to different finetuning components. We reported the performance under three different settings, 1) feature map (decomposition) only, 2) MLP decoder (LoRA) only, and 3) our method (both). Our PEFT method is used in experiments. As shown in Tab.~\ref{tab:ablation_component} and Fig.~\ref{fig:ablation_component}, performance is sigificantly dropped during the finetuning stage if any component is left out, especially the feature map.

\begin{table}[!h]
\centering
\renewcommand{\arraystretch}{1.2}
\resizebox{\columnwidth}{!}{
\begin{tabular}{l|ccccccccc}
\multicolumn{1}{c|}{\multirow{3}{*}{Method}} & \multicolumn{9}{c}{Iterations}                                                                                \\ \cline{2-10} 
\multicolumn{1}{c|}{}                        & \multicolumn{3}{c|}{500}                & \multicolumn{3}{c|}{1000}               & \multicolumn{3}{c}{2000} \\ \cline{2-10} 
\multicolumn{1}{c|}{}                        & PSNR & SSIM & \multicolumn{1}{c|}{MSIM} & PSNR & SSIM & \multicolumn{1}{c|}{MSIM} & PSNR   & SSIM   & MSIM   \\ \hline
Feat. only                                       &  25.29    & 0.886     & \multicolumn{1}{c|}{0.938}     & 26.49     & 0.901     & \multicolumn{1}{c|}{0.952}     & 27.50       & 0.911       & 0.960      \\
MLP only                              & 25.45     & 0.890     & \multicolumn{1}{c|}{0.934}     & 26.23     & 0.897     & \multicolumn{1}{c|}{0.941}     & 27.05       & 0.904       & 0.948       \\
\textbf{Ours}                                       & 27.13     & 0.905     & \multicolumn{1}{c|}{0.952}     & 28.79     & 0.923     & \multicolumn{1}{c|}{0.967}     & 29.74       & 0.934       & 0.975  
\end{tabular}}
\caption{Quantitative results on different finetuning component on objaverse dataset, `MISM' is MSSSIM.}
\label{tab:ablation_component}
\end{table}

\section{Additional Experiments}
\label{sec:additional_experiments}

\subsection{Comparison to Meta-initialization}
\label{sub:compare_meta}
We accessed our initialization and test time optimization with MetaINR~\cite{tancik2021learnedinit} on class-specific ShapeNet~\cite{chang2015shapenet, sitzmann2019scene} dataset. We used `Car' and `Chair' datasets that have a resolution of 128$\times$128. For a fair comparison, we set the same training configurations only except for the meta learning method. Specifically, 24 images are used to train the base model, and the same number of images are used for the test time optimization, then remaining views are evaluated for the metrics. Tab.~\ref{table:meta} reports the quantitative results, and our method shows faster convergence than the baseline from initialization even in the parameter efficient setting. 

\begin{table}[!hbt]
\renewcommand{\arraystretch}{1.2}
\resizebox{\columnwidth}{!}{
\begin{tabular}{c|c|cccccccc}
                        &                           & \multicolumn{8}{c}{Iterations}                                                                                                      \\ 
                        &                           & \multicolumn{2}{c}{0}                                         & \multicolumn{2}{c}{50}                                                 & \multicolumn{2}{c}{200}                                       & \multicolumn{2}{c}{1000}                                      \\ \cline{3-10} 
\multirow{-2}{*}{Data}  & \multirow{-3}{*}{Method}  & \multicolumn{1}{l}{PSNR}      & \multicolumn{1}{l}{SSIM}      & \multicolumn{1}{l}{PSNR}      & \multicolumn{1}{l}{SSIM}     & \multicolumn{1}{l}{PSNR}      & \multicolumn{1}{l}{SSIM}      & \multicolumn{1}{l}{PSNR}      & \multicolumn{1}{l}{SSIM}      \\ \hline
                        & MetaINR                   & 19.21                         & 0.846                         & 24.09                         & \cellcolor[HTML]{FFCCC9}0.904                         & 25.24                         & \cellcolor[HTML]{FFCCC9}0.917                         & 26.96                         & \cellcolor[HTML]{FFCCC9}0.935                         \\
\multirow{-2}{*}{Car}   & \textbf{Ours (PEFT)} & \cellcolor[HTML]{FFCCC9}23.34                         & \cellcolor[HTML]{FFCCC9}0.892                         & \cellcolor[HTML]{FFCCC9}24.53                         & \cellcolor[HTML]{FFCCC9}0.904                                             & \cellcolor[HTML]{FFCCC9}25.60                         & 0.915                         & \cellcolor[HTML]{FFCCC9}28.01 & \cellcolor[HTML]{FFCCC9}0.935                         \\ \hline
                        & MetaINR                   & 13.06                         & 0.603                         & 20.93                         & 0.816                       & \cellcolor[HTML]{FFCCC9}22.90                         & \cellcolor[HTML]{FFCCC9}0.859                         & 24.96                         & 0.889                         \\
\multirow{-2}{*}{Chair} & \textbf{Ours (PEFT)} & \cellcolor[HTML]{FFCCC9}20.24                         & \cellcolor[HTML]{FFCCC9}0.803                         & \cellcolor[HTML]{FFCCC9}21.48                         & \cellcolor[HTML]{FFCCC9}0.827                         & 22.84                         & 0.855                         & \cellcolor[HTML]{FFCCC9}25.83                         & \cellcolor[HTML]{FFCCC9}0.902                        
\end{tabular}}
\caption{Test time optimization comparison with MetaINR.}
\label{table:meta}
\end{table}

\begin{table*}[!h]
\renewcommand{\arraystretch}{1.1}
\resizebox{\linewidth}{!}{%
\begin{tabular}{l|l|ccccccccccccccc} \hline
\multicolumn{1}{c|}{}                       & \multicolumn{1}{c|}{}                         & \multicolumn{14}{c}{Iterations}                         \\ \cline{3-17} 
\multicolumn{1}{c|}{}                       & \multicolumn{1}{c|}{}                         & \multicolumn{3}{c|}{0}                                                                                             & \multicolumn{4}{c|}{500}                                                                                           & \multicolumn{4}{c|}{1000}                                                                                          & \multicolumn{4}{c}{2000}                                                                                      \\ \cline{3-17} 
\multicolumn{1}{c|}{\multirow{-3}{*}{Data}} & \multicolumn{1}{c|}{\multirow{-3}{*}{Method}} & \multicolumn{1}{l}{PSNR}      & \multicolumn{1}{l}{SSIM}      & \multicolumn{1}{l|}{MSIM}                          & \multicolumn{1}{l}{PSNR}      & \multicolumn{1}{l}{SSIM}      & \multicolumn{1}{l}{MSIM}   & \multicolumn{1}{l|}{SIZE}                       & \multicolumn{1}{l}{PSNR}      & \multicolumn{1}{l}{SSIM}      & \multicolumn{1}{l}{MSIM} & \multicolumn{1}{l|}{SIZE}                         & \multicolumn{1}{l}{PSNR}      & \multicolumn{1}{l}{SSIM}      & \multicolumn{1}{l}{MSIM} & \multicolumn{1}{l}{SIZE}                           \\ \hline
                                            & Triplanes                                     & 8.68                          & 0.713                         & \multicolumn{1}{c|}{0.266}                         & 22.57                         & 0.862           & 0.898              & \multicolumn{1}{c|}{8.143}                         & 25.51                         & 0.916                   & 0.955      & \multicolumn{1}{c|}{8.143}                         & 26.31                         & 0.934                & 0.966         & \multicolumn{1}{c}{8.143}                           \\
                                            & \textbf{Ours (PEFT)}                     & \cellcolor[HTML]{FFCCC9}24.98 & \cellcolor[HTML]{FFCCC9}0.910 & \multicolumn{1}{c|}{\cellcolor[HTML]{FFCCC9}0.951} & \cellcolor[HTML]{FFCCC9}25.86 & \cellcolor[HTML]{FFCCC9}0.915 & \cellcolor[HTML]{FFCCC9}0.957 & \multicolumn{1}{c|}{1.032} & \cellcolor[HTML]{FFCCC9}26.77 & \cellcolor[HTML]{FFCCC9}0.921 & \cellcolor[HTML]{FFCCC9}0.963 & \multicolumn{1}{c|}{1.032} & \cellcolor[HTML]{FFCCC9}27.24 & \cellcolor[HTML]{FFCCC9}0.927 & \cellcolor[HTML]{FFCCC9}0.966 & \multicolumn{1}{l}{1.032}\\
\multirow{-3}{*}{Car}                     & \textbf{Ours (PEFT++)}                   & $\cdot$                         & $\cdot$                         & \multicolumn{1}{c|}{$\cdot$}                         & 25.84                         & \cellcolor[HTML]{FFCCC9}0.915                         & \cellcolor[HTML]{FFCCC9}0.957 & \multicolumn{1}{c|}{\cellcolor[HTML]{FFCCC9}0.206}                         & 26.72                         & 0.920                         & \cellcolor[HTML]{FFCCC9}0.963 & \multicolumn{1}{c|}{\cellcolor[HTML]{FFCCC9}0.188} & 27.18                         & 0.926 & \cellcolor[HTML]{FFCCC9}0.966 & \multicolumn{1}{l}{\cellcolor[HTML]{FFCCC9}0.157}  \\ \hline
                                            & Triplanes                                     & 8.75                         & 0.740                         & \multicolumn{1}{c|}{0.274}                         & 19.73                         & 0.854                         & 0.780  & \multicolumn{1}{c|}{8.143}                          & 25.06                         & 0.930                         & 0.943  & \multicolumn{1}{c|}{8.143}                         & 26.22                         & 0.946                         & 0.961 & \multicolumn{1}{l}{8.143}                            \\
                                            & \textbf{Ours (PEFT)}                     & \cellcolor[HTML]{FFCCC9}25.44 & \cellcolor[HTML]{FFCCC9}0.928 & \multicolumn{1}{c|}{\cellcolor[HTML]{FFCCC9}0.952} & \cellcolor[HTML]{FFCCC9}29.87 & \cellcolor[HTML]{FFCCC9}0.954 & \cellcolor[HTML]{FFCCC9}0.979 & \multicolumn{1}{c|}{1.033} & \cellcolor[HTML]{FFCCC9}31.71 & \cellcolor[HTML]{FFCCC9}0.966 & \cellcolor[HTML]{FFCCC9}0.986 & \multicolumn{1}{c|}{1.033} & \cellcolor[HTML]{FFCCC9}32.75 & \cellcolor[HTML]{FFCCC9}0.973 & \cellcolor[HTML]{FFCCC9}0.989 & \multicolumn{1}{l}{1.033}      \\
\multirow{-3}{*}{Chair}                       & \textbf{Ours (PEFT++)}                   & $\cdot$                         & $\cdot$                         & \multicolumn{1}{c|}{$\cdot$}                         & 29.77                         & 0.953                         & 0.979 & \multicolumn{1}{c|}{\cellcolor[HTML]{FFCCC9}0.206}                         & 31.54                         & 0.965                         & 0.986 & \multicolumn{1}{c|}{\cellcolor[HTML]{FFCCC9}0.188}                         & 32.65                         & 0.972                         & \cellcolor[HTML]{FFCCC9}0.989  & \multicolumn{1}{l}{\cellcolor[HTML]{FFCCC9}0.158}
\end{tabular}%
}
\caption{Quantitative results of the proposed methods evaluated on ShapeNet `Car' and `Chair' categoris. `PEFT++' denotes parameter efficient finetuning with entropy coding, `MSIM' is MSSSIM and `SIZE' is measured in MB scale.}
\label{table:quan_shape}
\end{table*}

\subsection{Category-specific view synthesis}
We conducted experiments on ShapeNet~\cite{chang2015shapenet} for category-specific novel view synthesis. Note that the expeirment in~\ref{sub:compare_meta}, we used 128$\times$128 resolution ShapeNet dataset, but here, we used 224$\times$224 resolution and sourced the training/testing split in DISN~\cite{xu2019disn}. To train our base model, we also choose random 16 input images and camera poses and predicted the remaining novel views, and tested (finetuned) with the same setting in main paper. Notable changes are the increased number of decoder layers from 6 to 8 with the same hidden dimension 64 and decreased codebook size from $K=8192$ to $K=2048$ with the same dimension $C'=32$.

\subsubsection{Benchmarks}
\label{sub:sub:benchmarks}
Tab.~\ref{table:quan_shape} shows the quantitative results on ShapeNet dataset, and our model achieved fast encoding from initial representations and outperforms the baseline on all metrics. We also report the component level memory breakdown in Tab.~\ref{table:compression_car} and~\ref{table:compression_chair} on each category. Same as the Objaverse experiment in the main paper, it is measured after 10k iterations.

\begin{table}[h!]
\centering
\renewcommand{\arraystretch}{1.0}
\resizebox{1.0\linewidth}{!}{
\begin{tabular}{c|cccc}
\multirow{2}{*}{Component} & \multicolumn{4}{c}{Total size in MB (codes + finetuning deltas)}                                                                                                             \\ \cline{2-5} 
                           & \multicolumn{1}{c|}{Triplanes} &  \multicolumn{1}{c|}{PEFT} & \multicolumn{1}{c|}{PEFT++} & \multicolumn{1}{l}{W/O FT} \\ \hline
Codebook                   & \multicolumn{1}{c|}{.}        & \multicolumn{1}{c|}{0.013}         & \multicolumn{1}{c|}{0.013}      & 0.013                         \\
Feature                   & \multicolumn{1}{c|}{7.864}          & \multicolumn{1}{c|}{\cellcolor[HTML]{FFFC9E}0.984}   & \multicolumn{1}{c|}{\cellcolor[HTML]{FFCCC9}0.061}     & .                         \\
MLP                        & \multicolumn{1}{c|}{0.279}           & \multicolumn{1}{c|}{\cellcolor[HTML]{FFCCC9}0.035}   & \multicolumn{1}{c|}{\cellcolor[HTML]{FFCCC9}0.035}     & .                         \\ \hline
Total   & \multicolumn{1}{c|}{8.143}          & \multicolumn{1}{c|}{\cellcolor[HTML]{FFFC9E}1.032}    & \multicolumn{1}{c|}{\cellcolor[HTML]{FFCCC9}0.109}     & 0.013 
\end{tabular}}
\caption{Memory footprint in component level (`Car').}
\label{table:compression_car}
\end{table}

\begin{table}[h!]
\centering
\renewcommand{\arraystretch}{1.0}
\resizebox{1.0\linewidth}{!}{
\begin{tabular}{c|cccc}
\multirow{2}{*}{Component} & \multicolumn{4}{c}{Total size in MB (codes + finetuning deltas)}                                                                                                             \\ \cline{2-5} 
                           & \multicolumn{1}{c|}{Triplanes} &  \multicolumn{1}{c|}{PEFT} & \multicolumn{1}{c|}{PEFT++} & \multicolumn{1}{l}{W/O FT} \\ \hline
Codebook                   & \multicolumn{1}{c|}{.}        & \multicolumn{1}{c|}{0.013}         & \multicolumn{1}{c|}{0.013}      & 0.013                         \\
Feature                   & \multicolumn{1}{c|}{7.864}          & \multicolumn{1}{c|}{\cellcolor[HTML]{FFFC9E}0.984}   & \multicolumn{1}{c|}{\cellcolor[HTML]{FFCCC9}0.054}     & .                         \\
MLP                        & \multicolumn{1}{c|}{0.279}           & \multicolumn{1}{c|}{\cellcolor[HTML]{FFCCC9}0.035}   & \multicolumn{1}{c|}{\cellcolor[HTML]{FFCCC9}0.035}     & .                         \\ \hline
Total   & \multicolumn{1}{c|}{8.143}          & \multicolumn{1}{c|}{\cellcolor[HTML]{FFFC9E}1.032}    & \multicolumn{1}{c|}{\cellcolor[HTML]{FFCCC9}0.102}     & 0.013 
\end{tabular}}
\caption{Memory footprint in component level (`Chair').}
\label{table:compression_chair}
\end{table}

\section{Reproducibility}

\subsubsection{2D feature extraction}
We used a pre-trained visual transformer, DINO~\cite{caron2021dino}, as a feature extractor. It takes $256 \times 256$ images as input and produces feature tokens of dimension 768 from each image patch. We dropped out the \texttt{[CLS]} token, and used deconvolution operator to unpatchify the feature tokens.

\subsubsection{3D feature construction and compression}
\begin{figure*}[]
  \centering
  \includegraphics[width=0.81\linewidth]{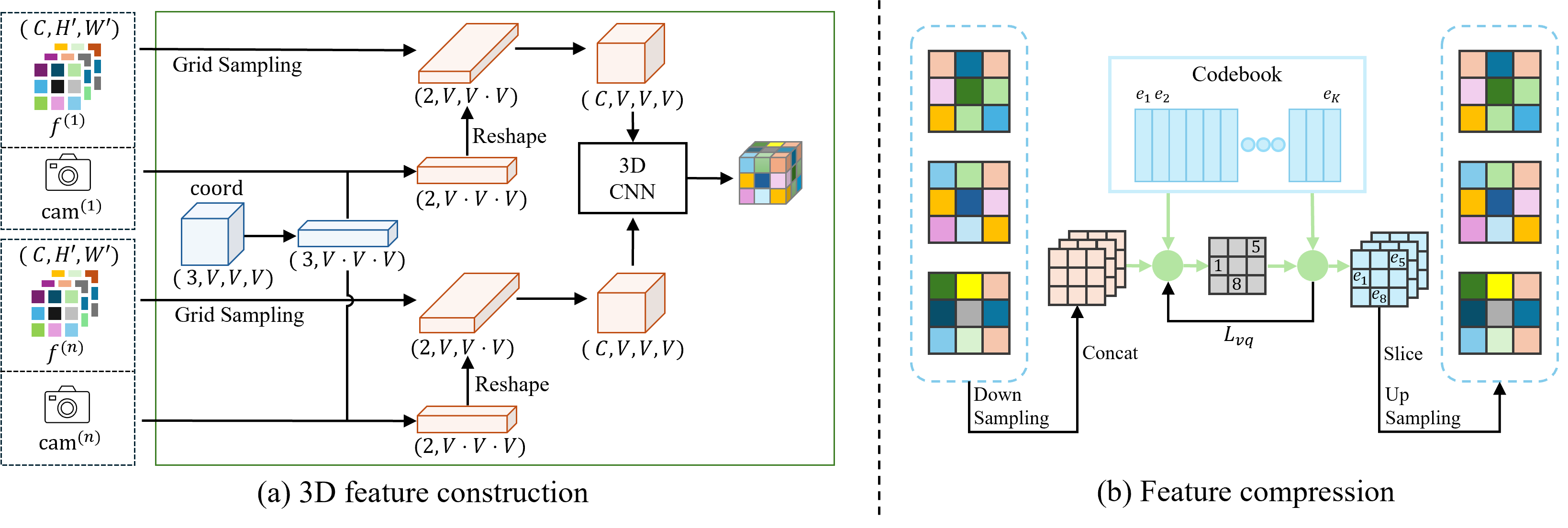}
  \caption{3D feature construction and compression architecture}
  \label{fig:unpro_vq}
\end{figure*}

We construct the 3D feature from multi-view input images. As shown in Fig.~\ref{fig:unpro_vq}(a), using extracted 2D feature maps from DINO~\cite{caron2021dino}, we unproject them to 3D coordinate tensor followed by the light-weight 3D CNN to aggregate 3D features. Next, we transform the 3D feature into three 2D features by pooling operation along each axis, and compress the three 2D feature maps using vector quantization module. The process is depicted in Fig.~\ref{fig:unpro_vq}(b).

\subsubsection{3D-aware 2D convolution}
\begin{figure*}[]
  \centering
  \includegraphics[width=0.81\linewidth]{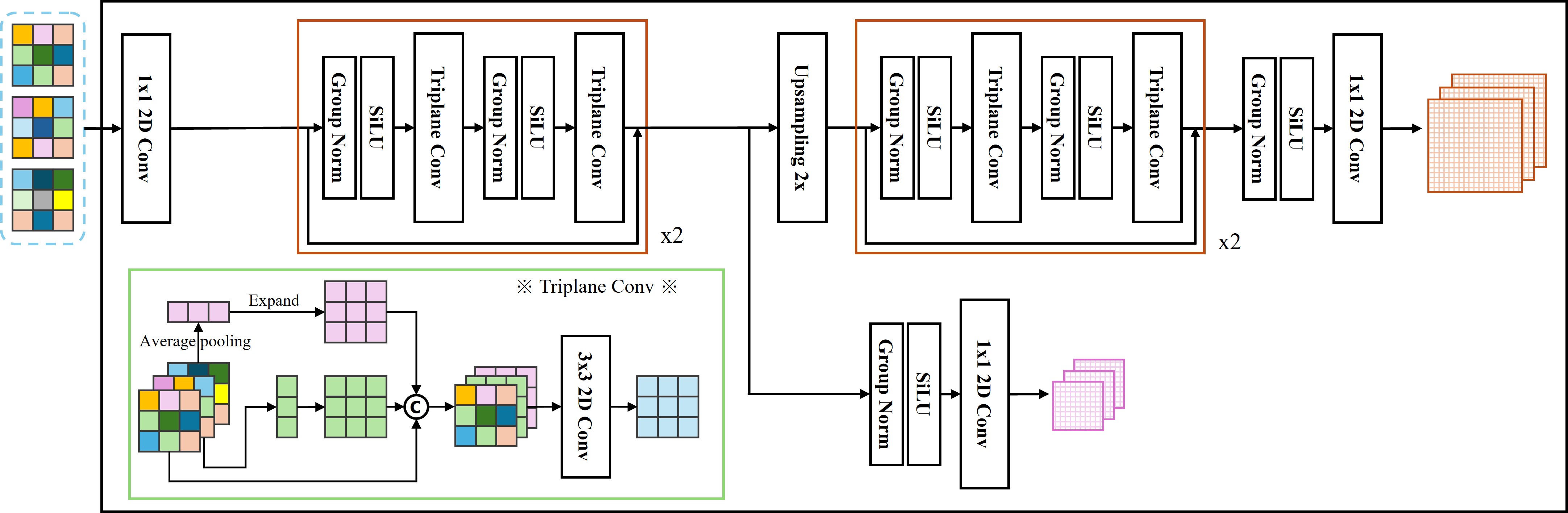}
  \caption{3d aware 2D convolution block.}
  \label{fig:3daware}
\end{figure*}

As briefly discussed in the main paper, we use a hierarchical 3D-aware 2D convolutional block to process the triplane features while respecting their 3D relationship. To compute for new $XY$ plane while attending to all elements in $YZ$ and $XZ$ planes, we perform axis-wise average pooling to $YZ$ (along $Z$ axis) and $XZ$ (along $Z$ axis) planes, resulting in two feature vectors. Then, aggregated vectors are expanded to the original 2D dimension by duplicating along the axis, concatenated with $XY$ plane channel-wise, and we perform a usual 2D convolution. The same procedure is applied to $YZ$ and $XZ$ planes. The overall architecture is depicted in Fig.~\ref{fig:3daware}, and we generate multi-resolution triplanes in a hierarchical manner.

\subsubsection{ResNet style 2D convolution}
\begin{figure}[!hbt]
  \centering
  \includegraphics[width=\linewidth]{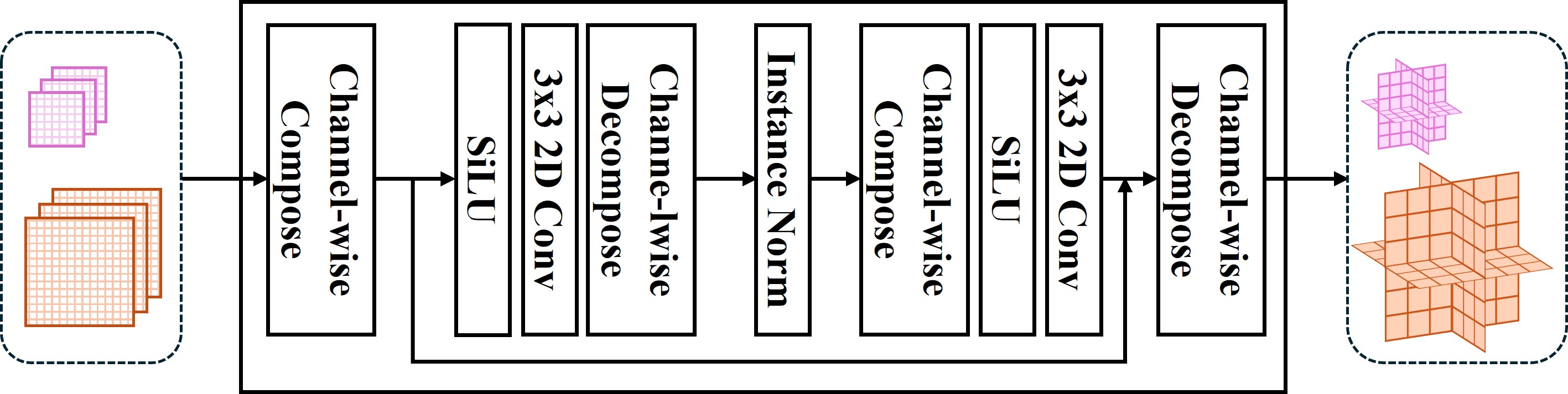}
  \caption{ResNet style 2D convolution block.}
  \label{fig:resstyle}
\end{figure}

We employ a ResNet style 2D convolution block to each multi-resolution triplanes, generated from 3D-aware 2D convolution module. This architecture is illustrated in Fig.~\ref{fig:resstyle}, and we can interpret this procedure as refining the triplane features before feeding into the MLP decoder.

\subsubsection{Decoding method}
We used vanilla NeRF~\cite{nerf} decoding method in all experiments which uses coarse MLP and fine MLP, both with identical architectures. We first sample 64 points using stratified sampling and then generate important 64 points that are biased towards the relevant surface of the volume, given the output of coarse MLP. We have trained our model using proposal sampling strategy~\cite{barron2022mip, kplanes_2023}, but we found that the decoder weights became excessively small, leading to model destabilization even with minor variations in the finetuning stage. 

\subsubsection{Training details}

We trained our base model for 850,000 steps using a batch size of 4, which requires about 80 GB of VRAM on a single GPU. The LPIPS loss weight is 0.3 and TV loss weight is $1e^{-4}$. Applying Adam~\cite{kingma2014adam} optimizer, we used learning rate of $1e^{-4}$ for MLP decoders and $1e^{-5}$ for others with StepLR scheduler. The decay factor and steps are set to 0.1 and 450K, respectively. When finetuning, we set the feature map learning rate to $5e^{-3}$ and the decoder learning rate to $1e^{-3}$ with a cosine schedule~\cite{loshchilov2016sgdr}. With higher learning rate, ranged from $1e^{-2}$ to $3e^{-2}$, the performance increased rapidly in the beginning but showed some unstable cases afterwards. In the experiment presented in the main paper's rate-distortion curve, we optimized all models including our (PEFT++) method with 10k iterations. All experiments that contain the elapsed time were measured on a single NVIDIA (80G) H100 GPU. We implemented neural codec based on CompressAI library~\cite{begaint2020compressai}.

\subsubsection{Baseline configurations}
We used various 3D representation models across experiments. We elaborate the detailed configuration as follows:
\begin{itemize}
      \item \textbf{Triplanes}: We modified from K-Planes~\cite{kplanes_2023}, set the same feature map and decoder size with our base model across the experiments.
      \item \textbf{TensoRF~\cite{chen2022tensorf}}: We changed the number of voxels ($128^3$) to match our highest resolution ($128 \times 128$). Every model based on TensoRF, we changed the voxel numbers equally. We trained the model with 4k iterations in DTU experiment.
      \item \textbf{DVGO~\cite{sun2022dvgo}}: We also set the number of voxels ($128^3$), and trained in default settings. We changed the voxel numbers of the models that are based on DVGO. 
      \item \textbf{Plenoxels~\cite{fridovich2022plenoxels}}: We trained Plonoxels in default settings, but we found that it showed many floaters and blurry parts on the Objaverse dataset.
      \item \textbf{BiRF~\cite{shin2024binarynerf}}: We trained BiRF with official `small' configuration with varying number of features (2, 4, and 8).
      \item \textbf{3D-GS~\cite{kerbl20233dgs}}: In DTU experiment, We trained the model with 7k iterations in default settings.
      \item \textbf{GeoLRM~\cite{zhang2024geolrm}}: We tested GeoLRM on the GSO dataset with default setting. We downscaled the original rendered dataset from $512 \times 512$ to $256 \times 256$ to match our model's input resolution.
\end{itemize}

\begin{figure*}[!h]
  \centering
  \includegraphics[width=\linewidth]{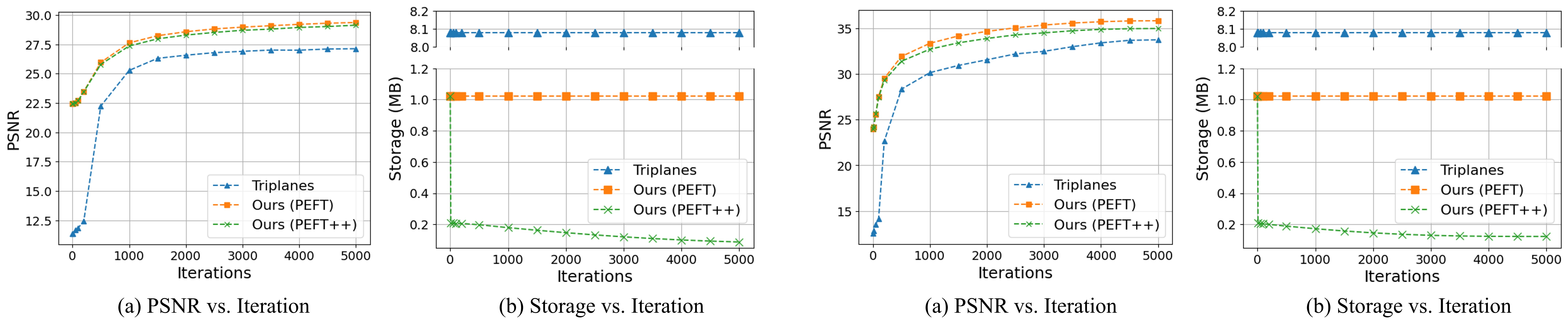}
  \caption{Comparison of optimization speed and compression degrees. Left two figures depict the results on the Objaverse dataset, and right two figures represent the results on the GSO dataset.}
  \label{fig:obj_gso}
\end{figure*}

\begin{figure*}[!h]
  \centering
  \includegraphics[width=\linewidth]{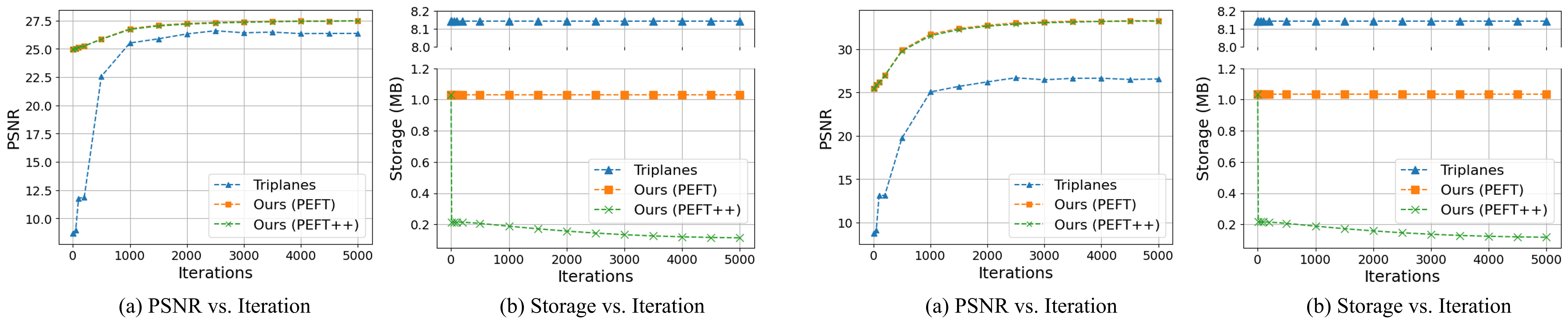}
  \caption{Comparison of optimization speed and compression degrees. Left two figures depict the results on the ShapeNet `Car' dataset, and right two figures represent the results on the ShapeNet `Chair' dataset.}
  \label{fig:car_chair}
\end{figure*}

\begin{figure*}[!h]
  \centering
  \includegraphics[width=0.9\linewidth]{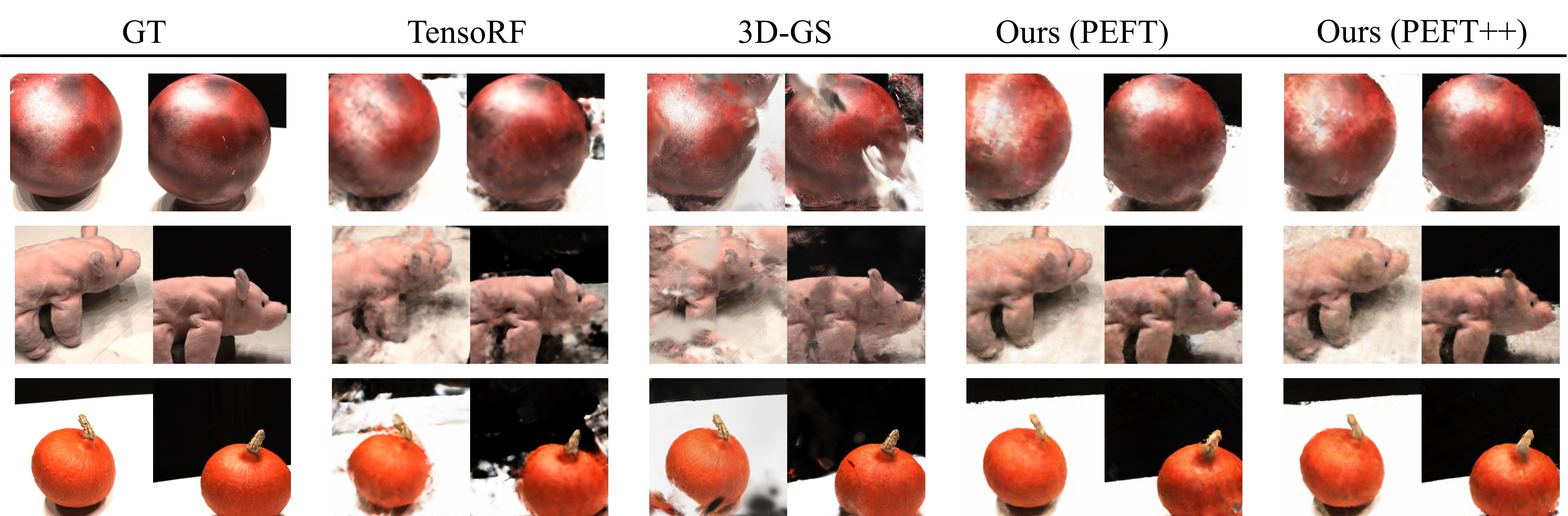}
  \caption{Novel view synthesis on the DTU dataset}
  \label{fig:dtu_sub}
\end{figure*}

\section{Limitations and Future Works}
\label{sec:limitation}
While CodecNeRF demonstrates promising performance in terms of fast encoding speed and compression ratio, it is important to acknowledge that the current framework still possesses limitations. First, further technical advancements are essential to encode more complicated scenes and objects, such as large-scale scenes (e.g., Mip-NeRF 360 datasets). Block-wise or hierarchical codings are promising directions to be explored, and training on large 3D scenes or videos could enhance the adaptability of CodecNeRF for such scenarios. Second, to support other NeRF representations, including instant NGP or 3D Gaussian Splatting, it will require modifications to the current architecture and training algorithms, potentially involving a point-based neural encoder and decoder. To further improve the rendering quality and encoding speed, we may consider investigating the utilization of larger encoder and decoder architectures and incorporating learned 2D priors~\cite{rombach2022ldm, radford2021clip} as a form of supervision. Lastly, we can utilize advanced techniques in neural codecs or weight-pruning methods to optimize compression performance.

\section{Additional Results}
\label{sec:additional_results}
\begin{figure*}[!h]
  \centering
  \includegraphics[width=\linewidth]{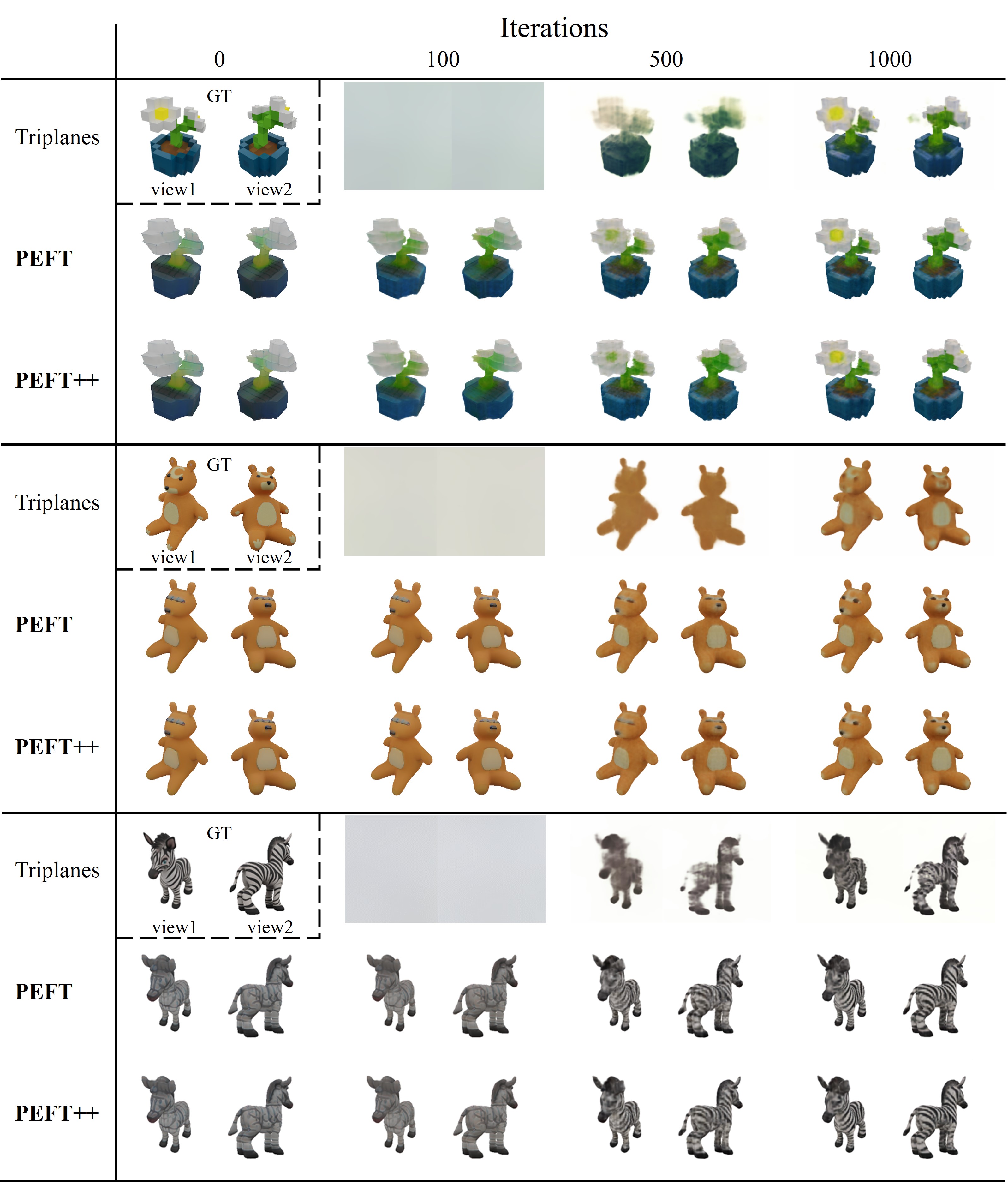}
  \caption{Novel view synthesis on the Objaverse dataset}
  \label{fig:sub1}
\end{figure*}

\begin{figure*}[!h]
  \centering
  \includegraphics[width=\linewidth]{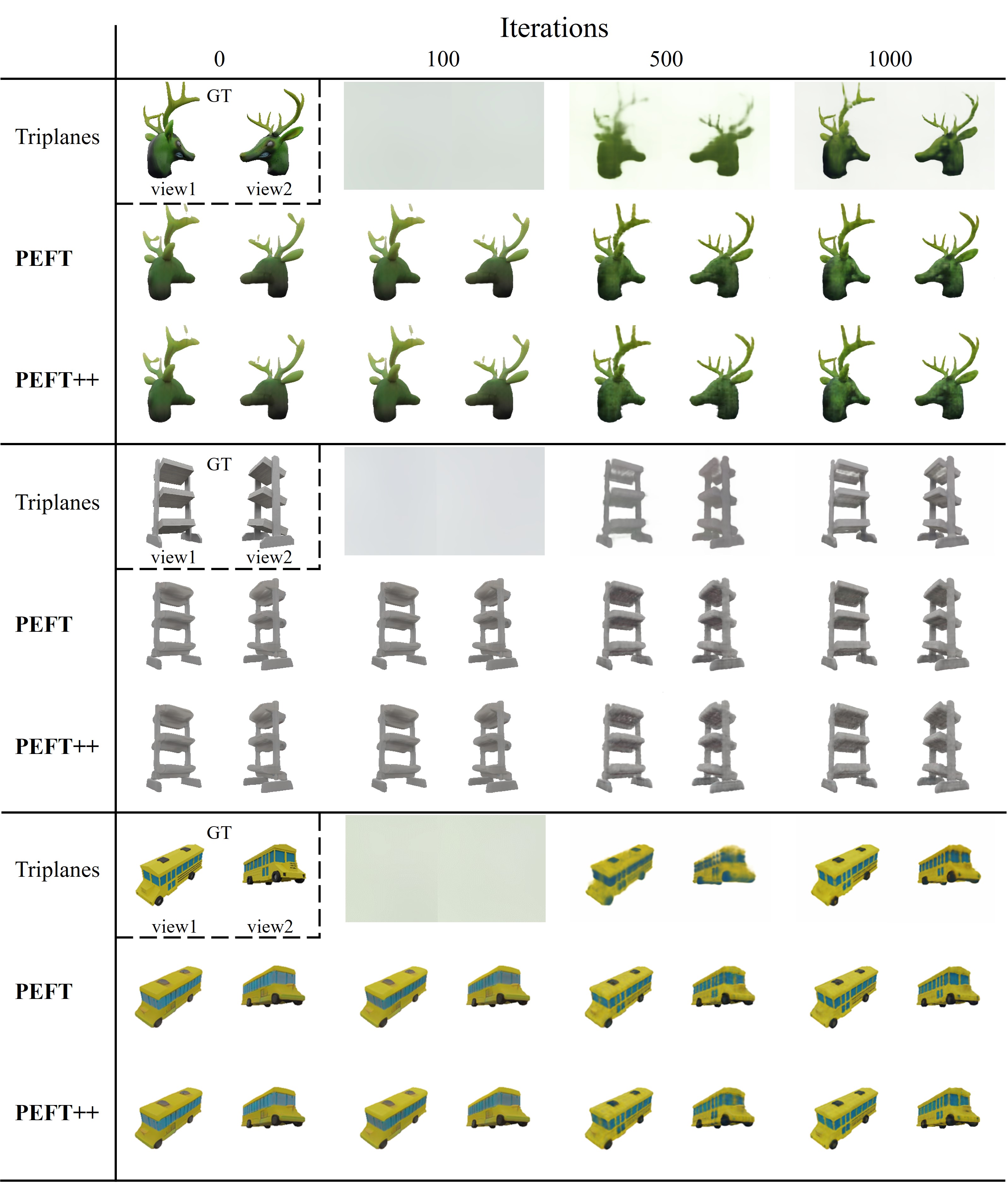}
  \caption{Novel view synthesis on the Objaverse dataset}
  \label{fig:sub2}
\end{figure*}

\begin{figure*}[!h]
  \centering
  \includegraphics[width=\linewidth]{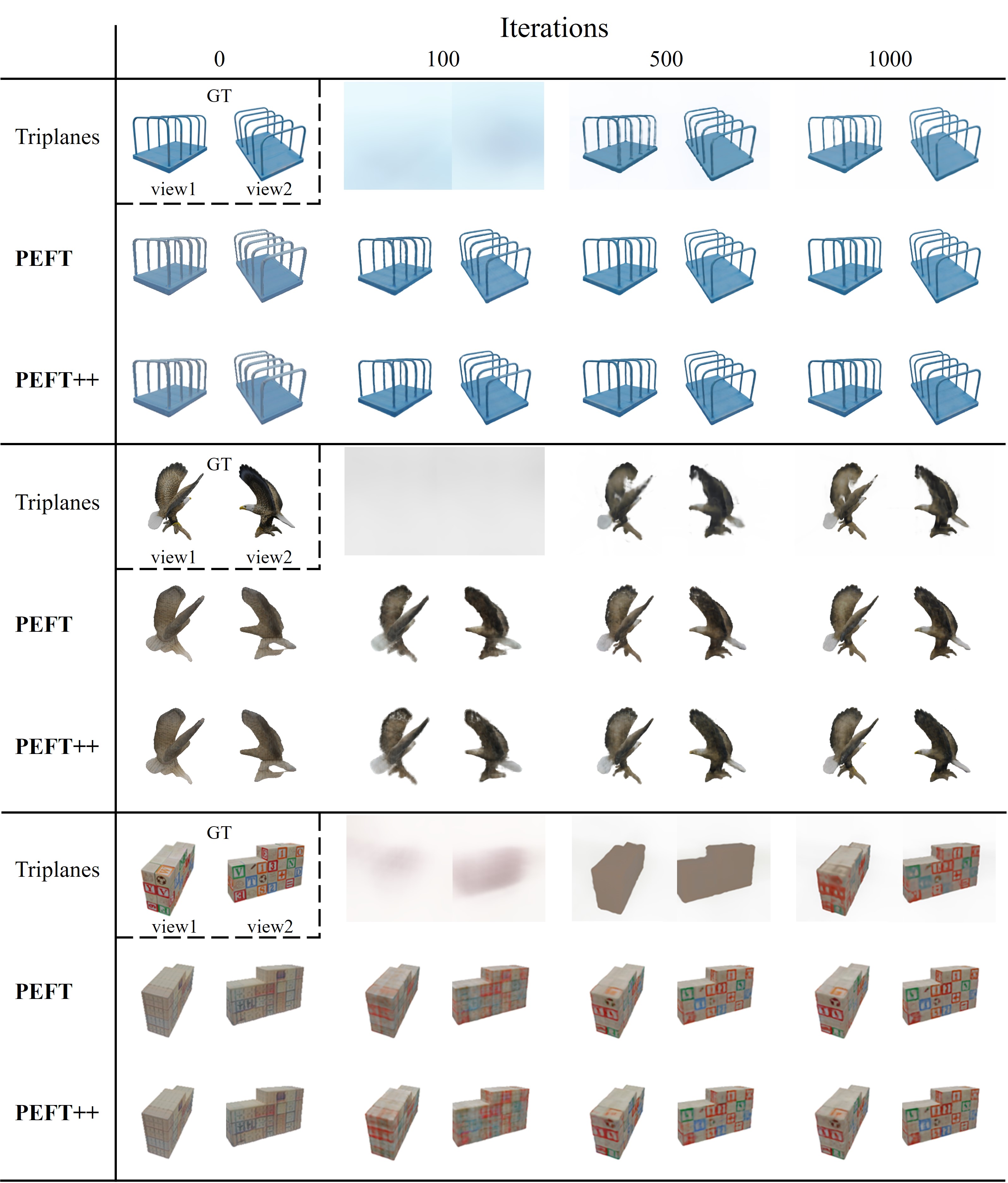}
  \caption{Novel view synthesis on the GSO dataset}
  \label{fig:sub3}
\end{figure*}

\begin{figure*}[!h]
  \centering
  \includegraphics[width=\linewidth]{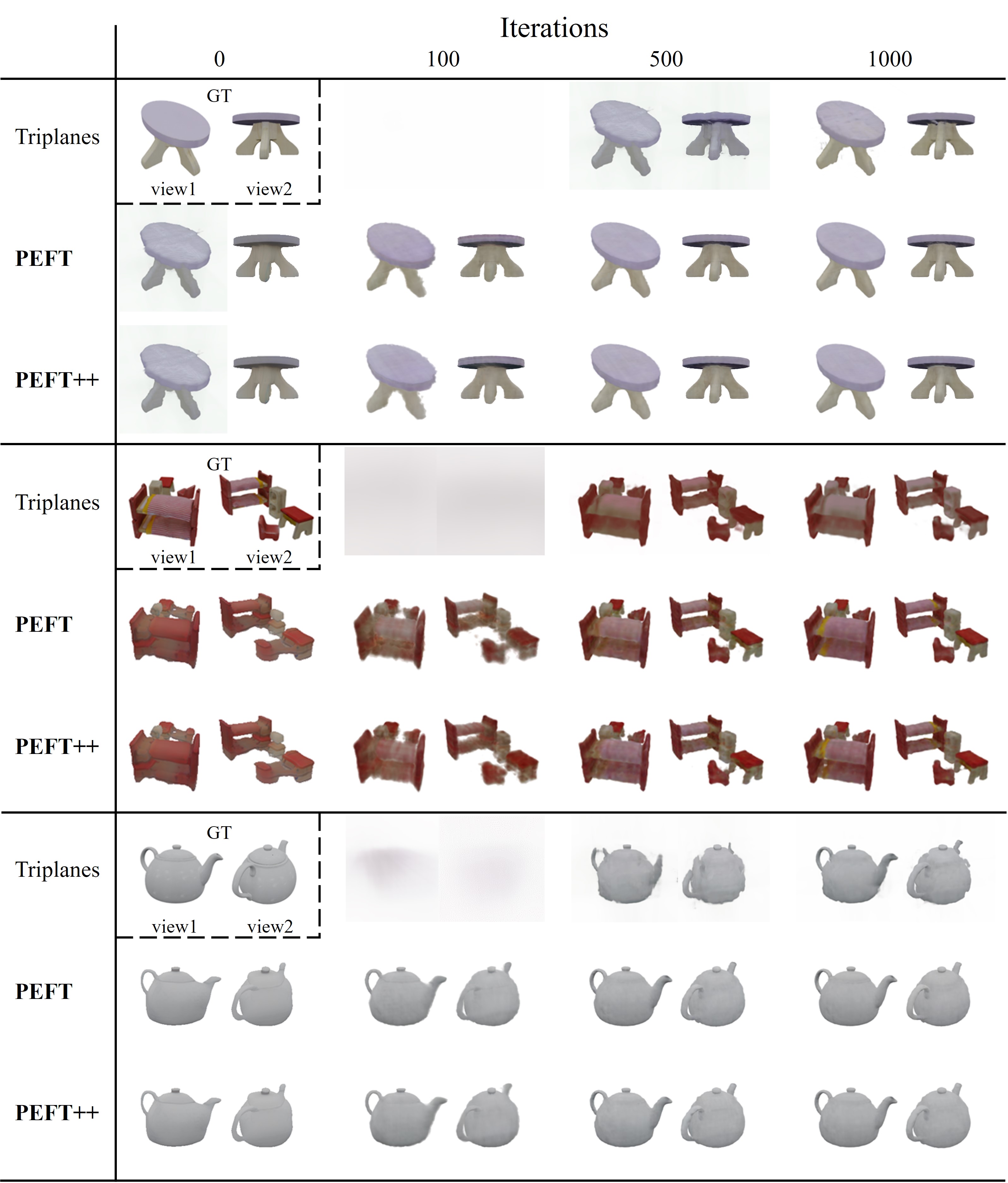}
  \caption{Novel view synthesis on the GSO dataset}
  \label{fig:sub4}
\end{figure*}

\begin{figure*}[!h]
  \centering
  \includegraphics[width=\linewidth]{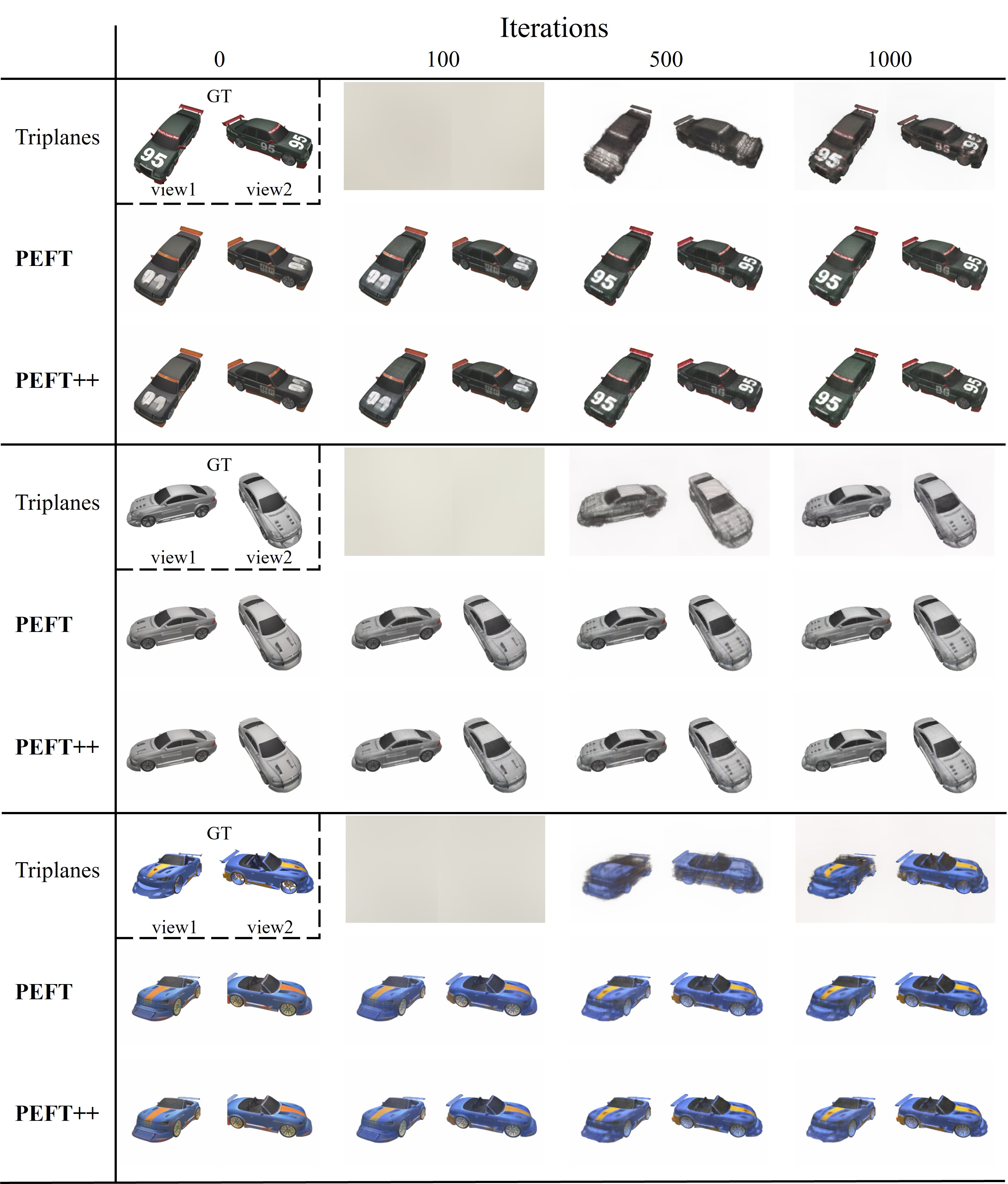}
  \caption{Novel view synthesis on the ShapeNet `Car' dataset}
  \label{fig:sub5}
\end{figure*}

\begin{figure*}[!h]
  \centering
  \includegraphics[width=\linewidth]{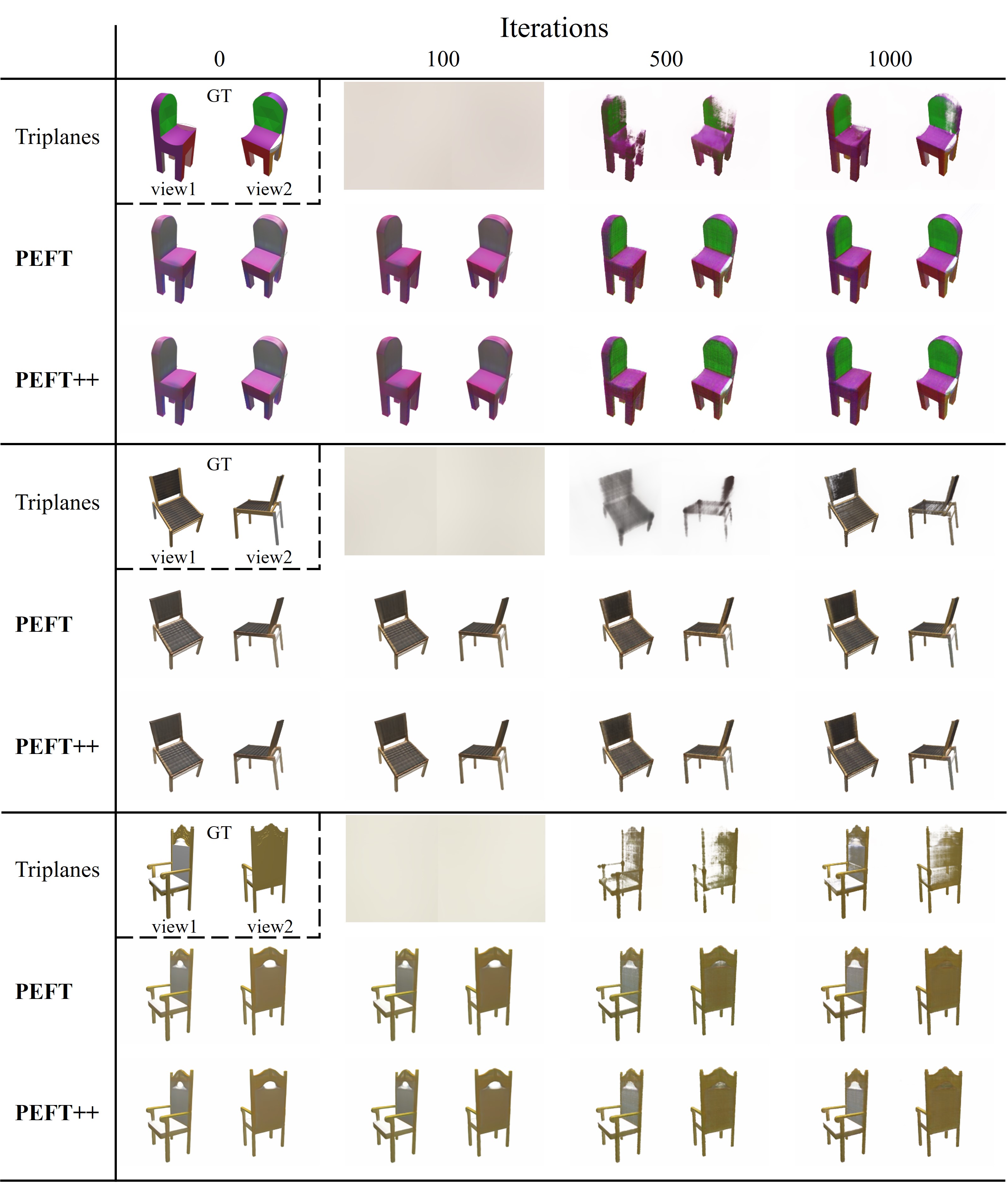}
  \caption{Novel view synthesis on the ShapeNet `Chair' dataset}
  \label{fig:sub6}
\end{figure*}

Fig.~\ref{fig:obj_gso} and~\ref{fig:car_chair} depicts the optimization and compression progress across the iterations in a quantitative manner. From Fig.~\ref{fig:dtu_sub} to Fig.~\ref{fig:sub6}, we shows the qualitative results for each of the datasets included.

\bibliography{aaai25}

\end{document}